\documentclass[10pt,journal,compsoc]{IEEEtran}

\usepackage{times}
\usepackage{epsfig}
\usepackage{graphicx}
\usepackage{amsmath}
\usepackage{amssymb}
\usepackage{url}
\usepackage{amsfonts}
\usepackage{xcolor}
\usepackage{bm}
\usepackage{multirow}
\usepackage{colortbl}
\usepackage{hhline}
\usepackage{pifont}
\usepackage{mathtools}
\usepackage{wrapfig}
\usepackage{cite}
\usepackage[caption=false]{subfig}
\usepackage{ragged2e}
\usepackage{makecell}
\usepackage[pagebackref=true,breaklinks=true,colorlinks,bookmarks=false]{hyperref}

\usepackage{booktabs}

\graphicspath{{./Imgs/}}
\DeclareGraphicsExtensions{.pdf,.jpg,.png}

\newcommand{\cmark}{\ding{51}}
\newcommand{\xmark}{\ding{55}}
\def\ourmodel{IOCFormer}
\def\ourmodelDepth{IOCFormer-D}
\def\ourdataset{IOCfish5K}
\def\ourdatasetDepth{IOCfish5K-D}
\newcommand{\myPara}[1]{\vspace{.01in}\noindent\textbf{#1}\quad}
\definecolor{mygray}{gray}{.92}

\newcommand{\figref}[1]{Fig.~\ref{#1}}

\newcommand{\secref}[1]{\S\ref{#1}}

\newcommand{\equref}[1]{Eq.~(\ref{#1})}

\def\ie{\emph{i.e.}}
\def\eg{\emph{e.g.}}

\def\etc{\emph{etc}}

\begin{document}

%%%%%%%%% TITLE
\title{RGB-D Indiscernible Object Counting in Underwater Scenes}

\markboth{IEEE TRANSACTIONS ON PATTERN ANALYSIS AND MACHINE INTELLIGENCE}%
{Indiscernible Object Counting in Underwater Scenes}

\author{
  Guolei Sun, Xiaogang Cheng, Zhaochong An, Xiaokang Wang, Yun Liu$^\dagger$, Deng-Ping Fan$^\dagger$, Ming-Ming Cheng, and Luc Van Gool
  \IEEEcompsocitemizethanks{%
    \IEEEcompsocthanksitem G. Sun and L. V. Gool are with the Computer Vision Lab, ETH Zurich, Zurich 8092, Switzerland.
    \IEEEcompsocthanksitem Z. An is with the Pioneer Centre for Artificial Intelligence, University of Copenhagen, Copenhagen 1350, Denmark.
    \IEEEcompsocthanksitem X. Cheng and X. Wang are with the School of Telecommunications and Information Engineering, Nanjing University of Posts and Telecommunications, Nanjing 210003, Jiangsu Province, China.
    \IEEEcompsocthanksitem Y. Liu, D.P. Fan, and M.M. Cheng are with the College of Computer Science, Nankai University, Tianjin 300350, China. 
    \IEEEcompsocthanksitem A preliminary version of this work has been published on CVPR 2023~\cite{sun2023indiscernible}.
    \IEEEcompsocthanksitem Corresponding authors: Yun Liu (liuyun@nankai.edu.cn) and Deng-Ping Fan (fdp@nankai.edu.cn)
  }
}

\IEEEtitleabstractindextext{%
%%%%%%%%% ABSTRACT
\begin{abstract} \justifying
Recently, indiscernible/camouflaged scene understanding has attracted lots of research attention in the vision community.
We further advance the frontier of this field by systematically studying a new challenge named \textit{indiscernible object counting} (\textbf{IOC}), the goal of which is to count objects that are blended with respect to their surroundings. Due to a lack of appropriate IOC datasets, we present a large-scale dataset \textbf{\ourdataset}~which contains a total of 5,637 high-resolution images and 659,024 annotated center points. Our dataset consists of a large number of indiscernible objects (mainly fish) in underwater scenes, making the annotation process all the more challenging. \ourdataset~is superior to existing datasets with indiscernible scenes because of its larger scale, higher image resolutions, more annotations, and denser scenes. All these aspects make it the most challenging dataset for IOC so far, supporting progress in this area. Benefiting from the recent advancements of depth estimation foundation models, we construct high-quality depth maps for \ourdataset~by generating pseudo labels using the \textit{Depth Anything V2} model. The RGB-D version of \ourdataset~is named \ourdatasetDepth. For benchmarking purposes on \ourdataset, we select 14 mainstream methods for object counting and carefully evaluate them. For multimodal \ourdatasetDepth, we evaluate other 4 popular multimodal counting methods.
Furthermore, we propose \textbf{\ourmodel}, a new strong baseline that combines density and regression branches in a unified framework and can effectively tackle object counting under concealed scenes. We also propose \textbf{\ourmodelDepth} to enable the effective usage of depth modality in helping detect and count objects hidden in their environments. Experiments show that \ourmodel~and \ourmodelDepth~achieve state-of-the-art scores on \ourdataset~and \ourdatasetDepth, respectively. The resources are publicly available at \href {https://github.com/GuoleiSun/Indiscernible-Object-Counting}{github.com/GuoleiSun/Indiscernible-Object-Counting}.
\end{abstract}

}

\maketitle

\IEEEdisplaynontitleabstractindextext
\IEEEpeerreviewmaketitle

%%%%%%%%% BODY TEXT
\section{Introduction}\label{sec:intro}
% \vspace{-3pt}

\begin{figure}[t!]
    \centering
    \includegraphics[width=.999\linewidth]{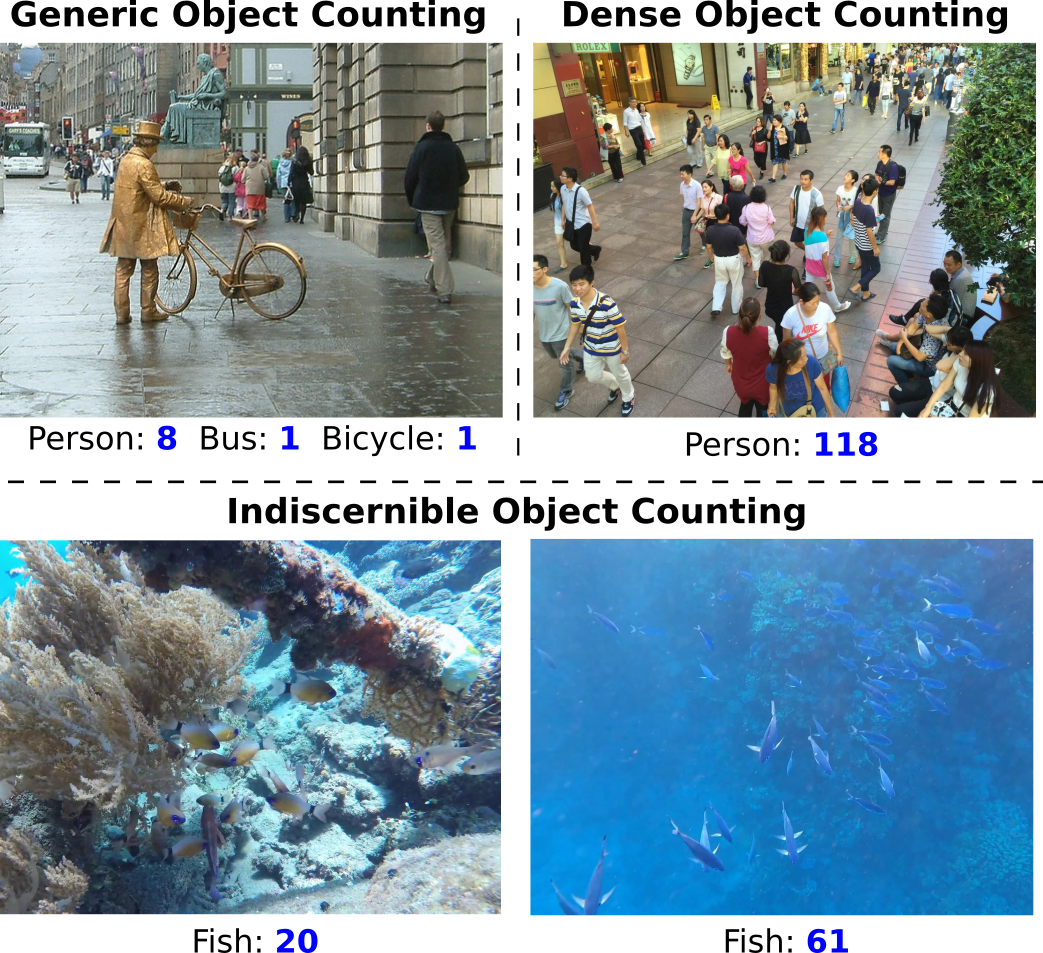}
    \caption{\textbf{Illustration of different counting tasks.} \textit{Top left}: Generic Object Counting (GOC), which counts objects of various classes in \textit{natural scenes}. \textit{Top right}: Dense Object Counting (DOC), which counts objects of a foreground class in \textit{scenes packed with instances}. \textit{Down}: Indiscernible Object Counting (IOC), which counts objects of a foreground class in \textit{indiscernible scenes}. Can you find all fishes in the given examples? For GOC, DOC, and IOC, the images shown are from PASCAL VOC~\cite{everingham2015pascal}, ShanghaiTech~\cite{zhang2016single}, and the new \ourdataset~dataset, respectively.}
    \label{fig:intro}
\end{figure}

Object counting -- to estimate the number of object instances in an image -- has always been an essential topic in computer vision. Understanding the counts of each category in a scene can be of vital importance for an intelligent agent to navigate in its environment. The task can be the end goal or can be an auxiliary step. As to the latter, counting objects has been proven to help instance segmentation~\cite{cholakkal2019object}, action localization~\cite{narayan20193c}, and pedestrian detection~\cite{xie2020count}. As to the former, it is a core algorithm in surveillance~\cite{wang2011automatic}, crowd monitoring~\cite{chan2008privacy}, wildlife conservation~\cite{norouzzadeh2018automatically}, diet patterns understanding~\cite{nguyen2022sibnet} and cell population analysis~\cite{alam2019machine}.

\begin{table*}[t!]
\centering
\caption{\textbf{Statistics of existing datasets for dense object counting (DOC) and indiscernible object counting (IOC).}}
\label{table:statis}
\small
\renewcommand{\arraystretch}{1.35}
\renewcommand{\tabcolsep}{1.5mm}
\resizebox{\linewidth}{!}{%
\begin{tabular}{l|c|c|c|r|c|c|rrrr|c}
\hline
\multirow{2}{*}{Dataset}  & \multirow{2}{*}{Year} & \multirow{2}{*}{\begin{tabular}[c]{@{}c@{}}Indiscernible\\ Scene\end{tabular}} & \multirow{2}{*}{Multimodal} & \multirow{2}{*}{\#Ann. IMG} & \multirow{2}{*}{Avg. Size} & \multirow{2}{*}{Free View} & \multicolumn{4}{c|}{Count Statistics}   & \multirow{2}{*}{Web} \\ \cline{8-11}
                         & &  &     &   &     &           & {Total}     & {Min} & {Ave}   & Max    &                       \\ 
                         \hline
UCSD~\cite{chan2008privacy}  &  2008  &       \xmark & \xmark  & 2,000                  & 158$\times$238                     &   \xmark    & {49,885}    & {11}  & {25}    & 46     &  \href{http://www.svcl.ucsd.edu/projects/peoplecnt/}{Link}             \\ 
%\hline
Mall~\cite{chen2012feature} & 2012  &       \xmark  &\xmark  & 2,000                  & 480$\times$640                     &    \xmark       & {62,325}    & {13}  & {31}    & 53     &  \href{http://personal.ie.cuhk.edu.hk/~ccloy/downloads_mall_dataset.html}{Link}              \\ 
%\hline
UCF\_CC\_50~\cite{idrees2013multi}   & 2013  &  \xmark & \xmark  & 50      & 2101$\times$2888                   &    \cmark       & {63,974}    & {94}  & {1,279} & 4,543  &   \href{http://crcv.ucf.edu/data/ucf-cc-50/}{Link}               \\ 
%\hline
WorldExpo'10~\cite{zhang2016data} & 2016  &       \xmark  & \xmark   & 3,980                  & 576$\times$720                     &    \xmark       & {199,923}   & {1}   & {50}    & 253    & \href{http://www.ee.cuhk.edu.hk/~xgwang/expo.html}{Link}                   \\ 
%\hline
ShanghaiTech B~\cite{zhang2016single}   &  2016    &       \xmark   & \xmark  & 716      & 768$\times$1024                    &    \xmark       & {88,488}    & {9}   & {123}   & 578    &  \href{https://github.com/desenzhou/ShanghaiTechDataset}{Link}                 \\ 
%\hline
ShanghaiTech A~\cite{zhang2016single}   &  2016 &     \xmark & \xmark  & 482      & 589$\times$868                     &    \cmark       & {241,677}   & {33}  & 501   & 3,139  &  \href{https://github.com/desenzhou/ShanghaiTechDataset}{Link}                 \\ 
%\hline
UCF-QNRF~\cite{idrees2018composition}   & 2018 &      \xmark & \xmark  & 1,535    & 2013$\times$2902    &    \cmark       & 1,251,642 & 49  & 815   & 12,865 &   \href{https://www.crcv.ucf.edu/data/ucf-qnrf/}{Link}                \\ 
%\hline
Crowd\_surv~\cite{yan2019perspective}   &    2019    &       \xmark  &  \xmark  & 13,945     & 840$\times$1342                    &    \xmark      & {386,513}   & {2}   & {35}    & 1420   & \href{https://ai.baidu.com/broad/introduction}{Link}                 \\ 
%\hline
GCC (synthetic)~\cite{wang2019learning}   & 2019 &   \xmark  &  \xmark   & 15,212     & 1080$\times$1920      &     \xmark      & 7,625,843 & 0   & 501   & 3,995  &  \href{https://mailnwpueducn-my.sharepoint.com/:f:/g/personal/gjy3035_mail_nwpu_edu_cn/Eo4L82dALJFDvUdy8rBm6B0BuQk6n5akJaN1WUF1BAeKUA?e=ge2cRg}{Link}                    \\ 
%\hline
JHU-CROWD++~\cite{sindagi2019pushing} & 2019  &  \xmark & \xmark  & 4,372   & 910$\times$1430 &     \cmark      & 1,515,005 & 0   & 346   & 25,791 &   \href{http://www.crowd-counting.com/}{Link}                    \\ 
%\hline
NWPU-Crowd~\cite{gao2020nwpu}   & 2020 &    \xmark  & \xmark  & 5,109    & 2191$\times$3209  &     \cmark      & 2,133,375 & 0   & 418   & 20,033 & \href{https://gjy3035.github.io/NWPU-Crowd-Sample-Code/}{Link}      \\ \hline
NC4K~\cite{yunqiu_cod21}  &  2021   &    \cmark & \xmark  &  4,121 &  530$\times$709  &  \cmark  & 4,584 &  1  & 1 & 8 & \href{https://github.com/JingZhang617/COD-Rank-Localize-and-Segment}{Link} \\ 
CAMO++~\cite{le2021camouflaged}  &     2021     &    \cmark & \xmark  & 5,500  &  N/A  &  \cmark  & 32,756 & N/A & 6   & N/A & \href{https://sites.google.com/view/ltnghia/research/camo_plus_plus}{Link} \\ 
COD~\cite{fan2021concealed} &     2022     &    \cmark & \xmark  & 5,066  &  737$\times$964  & \cmark   & 5,899 &  1  &  1   & 8 & \href{https://github.com/DengPingFan/SINet}{Link} \\ 
\textbf{\ourdataset~(Ours)}      &     2023     &   \cmark  & \xmark  & 5,637    & 1080$\times$1920   &    \cmark       & 659,024 & 0   & 117   & 2,371 &  \href{https://github.com/GuoleiSun/Indiscernible-Object-Counting}{Link}             \\ 
\textbf{\ourdatasetDepth~(Ours)}      &     2023     &   \cmark  & \cmark  & 5,637    & 1080$\times$1920   &    \cmark       & 659,024 & 0   & 117   & 2,371 &  \href{https://github.com/GuoleiSun/Indiscernible-Object-Counting}{Link}             \\ \hline
\end{tabular}
}
\end{table*}

Previous object counting research mainly followed two directions: generic/common object counting (GOC)~\cite{chattopadhyay2017counting,cholakkal2019object,laradji2018blobs,stahl2018divide} and dense object counting (DOC)~\cite{zhang2016single,idrees2018composition,li2018csrnet,sindagi2020jhu,onoro2016towards,lu2017tasselnet,song2021choose}. The difference between these two sub-tasks lies in the studied scenes, as shown in Fig.~\ref{fig:intro}. GOC tackles the problem of counting object(s) of various categories in natural/common scenes~\cite{chattopadhyay2017counting}, \ie, images from PASCAL VOC~\cite{everingham2015pascal} and COCO~\cite{lin2014microsoft}. The number of objects to be estimated is usually small, \ie, less than 10. DOC, on the other hand, mainly counts objects of a foreground class in crowded scenes. The estimated count can be hundreds or even tens of thousands. The counted objects are often persons (crowd counting)~\cite{li2018csrnet,yang2020weakly,liang2022end}, vehicles~\cite{onoro2016towards,hsieh2017drone} or plants~\cite{lu2017tasselnet}. Thanks to large-scale datasets~\cite{gao2020nwpu,zhang2016single,sindagi2019pushing,idrees2018composition,chen2012feature,everingham2015pascal} and deep convolutional neural networks (CNNs) trained on them, significant progress has been made both for GOC and DOC. However, to the best of our knowledge, there is no previous work on counting indiscernible objects.

% [Camouflaged scene understanding]
Under indiscernible/camouflaged scenes, foreground objects have a similar appearance, color, or texture to the background and are thus difficult to be detected with a traditional visual system. The phenomenon exists in both natural and artificial scenes~\cite{fan2020Camouflage,le2021camouflaged}. Hence, scene understanding for indiscernible scenes has attracted increasing attention since the appearance of some pioneering works~\cite{le2019anabranch,fan2020Camouflage}. Various tasks have been proposed and formalized: camouflaged object detection (COD)~\cite{fan2020Camouflage}, camouflaged instance segmentation (CIS)~\cite{le2021camouflaged} and video camouflaged object detection (VCOD)~\cite{lamdouar2020betrayed,cheng2022implicit}. However, no previous research has focused on counting objects in indiscernible scenes, which is an important aspect.

% [Camouflaged object counting]
In this paper, we study the new \textit{indiscernible object counting} (\textbf{IOC}) task, which focuses on counting foreground objects in indiscernible scenes. Fig.~\ref{fig:intro} illustrates this challenge. Tasks such as image classification~\cite{he2016deep,dosovitskiy2020image}, semantic segmentation~\cite{chen2017deeplab,liu2019auto} and instance segmentation~\cite{he2017mask,yolact-iccv2019} all owe their progress to the availability of large-scale datasets~\cite{deng2009imagenet,everingham2015pascal,lin2014microsoft}. Similarly, a high-quality dataset for IOC would facilitate its advancement. Although existing datasets~\cite{fan2020Camouflage,yunqiu_cod21,le2021camouflaged} with instance-level annotations can be used for IOC, they have the following limitations: 1) the total number of annotated objects in these datasets is limited, and image resolutions are low; 2) they only contain scenes with a small instance count; 3) the instance-level mask annotations can be converted to point supervision by computing mask centers, but the computed points do not necessarily fall inside the objects. 

\begin{figure*}[!t]
    \centering
    \includegraphics[width=.999\linewidth]{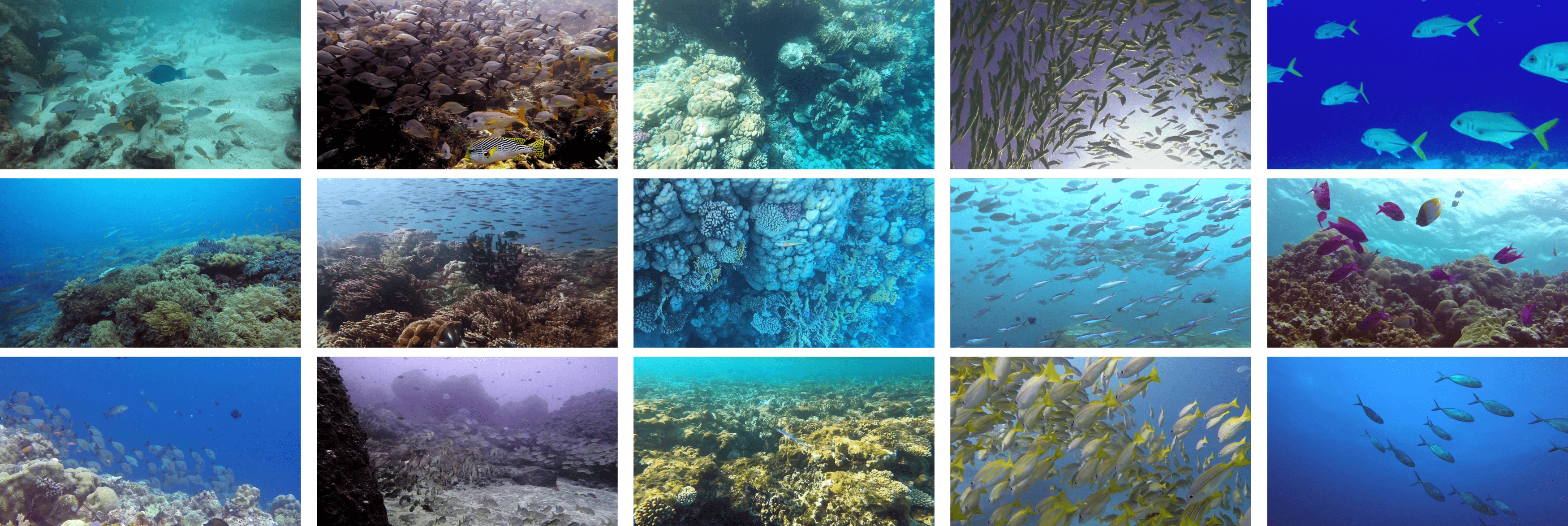}
     \caption{\textbf{Example images from the proposed \ourdataset.} From \textit{left} column to \textit{right} column: typical samples, indiscernible \& dense samples, indiscernible \& less dense samples, less indiscernible \& dense samples, less indiscernible \& less dense samples. 
     %\textit{Best viewed with zooming and in color}.
     }
    \label{fig:dataset_samples}
\end{figure*}

% [To facilitate the research on IOC, we propose a large-scale dataset]
To facilitate the research on IOC, we construct a large-scale dataset, \ourdataset. We collect 5,637 images with indiscernible scenes and annotate them with 659,024 center points. Compared with the existing datasets, the proposed \ourdataset~has several advantages: 1) it is the largest-scale dataset for IOC in terms of the number of images, image resolution, and total object count; 2) the images in \ourdataset~are carefully selected and contain diverse indiscernible scenes; 3) the point annotations are accurate and located at the center of each object. Our dataset is compared with existing DOC and IOC datasets in Table~\ref{table:statis}, and example images are shown in Fig.~\ref{fig:dataset_samples}. More details about building \ourdataset~is introduced in \secref{sec:camfish5k_dataset}.

% [introduce ourdataset with depth]
The ability to estimate depth is important to animal's visual system, allowing it to effectively perceive its environment. Depth cues play a crucial role in interpreting camouflaged scenes. Even when objects blend seamlessly with their surroundings due to similar appearance or poor light conditions, differences in depth can provide essential complementary information, making these objects distinguishable~\cite{yu2024exploring,wang2024depth}. Inspired by this, we construct a new RGB-D dataset \ourdatasetDepth, which contains high-quality depth maps for all the images in \ourdataset. To be more specific, we exploit a large vision foundation model, Depth Anything V2~\cite{yang2024depth2}, to generate pseudo depth maps for inputs in different sizes. We found that depth maps generated from small and large inputs have both advantages and disadvantages: small depth maps have better boundaries on large and close objects, while large depth maps are more accurate for small and distant objects. Therefore, we propose to merge pseudo depth maps output from different-size inputs, leading to high-quality pseudo depth maps. More details about constructing \ourdatasetDepth~is explained in \secref{sec:camfish5k_depth}.

% [In addition, we propose a method, to do]
Based on the proposed \ourdataset~dataset, we provide a systematic study on 14 mainstream baselines~\cite{zhang2016single,li2018csrnet,laradji2018blobs,liu2019context,liu2019crowd,ma2019bayesian,wan2020modeling,wang2020DMCount,song2021rethinking,zand2022multiscale,lin2022boosting,liang2022end}. We find that methods which perform well on existing DOC datasets do not necessarily preserve their competitiveness on our challenging dataset.
Hence, we propose a simple and effective approach named \ourmodel. Specifically, we combine the advantages of density-based~\cite{wang2020DMCount} and regression-based~\cite{liang2022end} counting approaches. The former can estimate the object density across the image, while the latter directly regresses the coordinates of points, which is straightforward and elegant. \ourmodel~contains two branches: density and regression. The density-aware features from the density branch help make indiscernible objects stand out through the proposed density-enhanced transformer encoder (DETE). Then the refined features are passed through a conventional transformer decoder, after which predicted object points are generated. Experiments show that \ourmodel~outperforms all considered algorithms, demonstrating its effectiveness on IOC. 

% [iocformer using depth]
Furthermore, to benchmark multimodal \ourdatasetDepth, we evaluate existing popular multimodal counting methods~\cite{Lian_2019_CVPR,liu2021cross,zhang2022spatio,meng2025multi}. However, these methods do not perform well since they are not specifically designed for object counting under indiscernible/camouflaged scenes. To mitigate this, we propose \ourmodelDepth~by extending \ourmodel~with the ability of fully exploiting depth modality. \ourmodelDepth~includes a tiny depth encoder to extract multi-scale depth features, which are then fused through multi-scale depth feature merging (MDFM) module. Next, RGB and depth features are integrated through the proposed cross-agent multimodal fusion (CAMF) module to generate comprehensive features, which are used for regression-based counting. Different modalities fully interact through exchanging agents, followed by agent attention~\cite{han2025agent}. Experiments indicate that \ourmodelDepth~achieves a new state-of-the-art performance on \ourdatasetDepth.

% [Our contributions are three-fold]
To summarize, our contributions are four-fold as follows:
\begin{itemize}
    % \vspace{-5.5pt}
    \item 
    We propose the new indiscernible object counting (IOC) task. To facilitate research on IOC, we contribute a large-scale dataset \ourdataset, containing 5,637 images and 659,024 accurate point labels. 
    % \vspace{-5.5pt}
    \item 
    We select 14 classical and high-performing approaches for object counting and evaluate them on the proposed \ourdataset~for benchmarking purposes. 
    % \vspace{-5.5pt}
    \item 
    We further introduce a RGB-D counting dataset \ourdatasetDepth~by generating high-quality depth maps for all images in \ourdataset. For benchmarking purposes, we also evaluate 4 popular multimodal counting methods on \ourdatasetDepth.
    \item 
    We propose a novel baseline, namely \ourmodel, which integrates density-based and regression-based methods in a unified framework. In addition, a novel density-based transformer encoder is proposed to gradually exploit density information from the density branch to help detect indiscernible objects. To fully exploit depth information, we extend \ourmodel~to \ourmodelDepth, outperforming all existing methods.
\end{itemize}

This paper is an extension of our previous work~\cite{sun2023indiscernible}. We significantly extend it in the following aspects. \textbf{(1)} Based on previously proposed unimodal dataset \ourdataset, we introduce a new RGB-D counting dataset \ourdatasetDepth, by generating high-quality depth maps for all the images using vision foundation model (\S\ref{sec:camfish5k_depth}). \textbf{(2)} For benchmarking purposes, we select 4 popular multimodal counting methods with open-source codes and systematically evaluate them on the \ourdatasetDepth. Both quantitative results and in-depth analyses for those methods are provided (\S\textcolor{red}{7}). \textbf{(3)} We propose a new RGB-D method \ourmodelDepth~for the IOC task, which extends \ourmodel~by exploiting rich and complementary depth information and achieves new state-of-the-art performance (\S\ref{sec:iocformer-d}). \textbf{(4)} We conduct extensive (ablation) experiments with \ourmodelDepth~and present both quantitative and qualitative results (\S\textcolor{red}{7}). \textbf{(5)} Compared to our conference version, we provide more details about benchmarking algorithms, more ablation studies, and more insightful discussions/analyses about our datasets as well as methods (\S\ref{sec:related-works}, \S\ref{sec:camfish5k_dataset}, \S\ref{sec:experiments}).

\section{Related Works}\label{sec:related-works}

\subsection{Generic Object Counting}
% \vspace{-3pt}
Generic/common object counting (GOC)~\cite{cholakkal2019object}, also referred to as everyday object counting~\cite{chattopadhyay2017counting}, is to count the number of object instances for various categories in natural scenes. 
% Given an input image, GOC methods could count number of instances for countable classes including \textit{fruits}, \textit{bus}, \textit{bicyble}, \textit{person} and so on. 
The popular benchmarks for GOC are PASCAL VOC~\cite{everingham2015pascal} and COCO~\cite{lin2014microsoft}. The task was first proposed and studied in the pioneering work~\cite{chattopadhyay2017counting}, which divided images into non-overlapping patches and predicted their counts by subitizing. LC~\cite{cholakkal2019object} used image-level count supervision to generate a density map for each class, improving counting performance and instance segmentation. RLC~\cite{cholakkal2020towards} further reduced the supervision by only requiring the count information for a subset of training classes rather than all classes. Differently, LCFCN~\cite{laradji2018blobs} exploited point-level supervision and output a single blob per object instance. 

\subsection{Dense Object Counting}
Dense object counting (DOC)~\cite{zhang2016single,idrees2018composition,xiong2019open,sindagi2020jhu,onoro2016towards,lu2017tasselnet,meng2021count,wen2021detection,shu2022crowd,xiong2022discrete,Cheng_2022_CVPR} counts the number of objects in dense scenarios. DOC contains tasks such as crowd counting~\cite{zhang2016single,idrees2018composition,sindagi2020jhu,gao2020nwpu,wang2021uniformity,9601215,9346018,xu2022autoscale,liu2020weighing,jiang2020attention}, vehicle counting~\cite{onoro2016towards,hsieh2017drone}, plant counting~\cite{lu2017tasselnet}, cell counting~\cite{alam2019machine} and penguin counting~\cite{arteta2016counting}. Among them, crowd counting, \ie, counting people, attracts the most attention. 
The popular benchmarks for crowd counting include ShanghaiTech~\cite{zhang2016single}, UCF-QNRF~\cite{idrees2018composition}, JHU-CROWD++~\cite{sindagi2020jhu}, NWPU-Crowd~\cite{gao2020nwpu} and Mall~\cite{chen2012feature}. For vehicle counting, researchers mainly use TRANCOS~\cite{onoro2016towards}, PUCPR+~\cite{hsieh2017drone}, and CAPRK~\cite{hsieh2017drone}. For DOC on other categories, the available datasets are MTC~\cite{lu2017tasselnet} for counting plants, CBC~\cite{alam2019machine} for counting cells, and Penguins~\cite{arteta2016counting} for counting penguins. DOC differs from GOC because DOC has far more objects to be counted and mainly focuses on one particular class. 

Previous DOC works can be divided into three groups based on the counting strategy: detection~\cite{liu2018decidenet,ge2009marked,li2008estimating,sam2020locate}, regression~\cite{chan2008privacy,chan2009bayesian,idrees2013multi,song2021rethinking,liang2022end}, and density map generation~\cite{li2018csrnet,liu2019context,shi2019counting,wang2020DMCount,lin2022boosting,sun2021boosting,liu2020adaptive}. Counting-by-detection methods first detect the objects and then count. Though intuitive, they are inferior in performance since detection performs unfavorably on crowded scenes.  Counting-by-regression methods either regress the global features to the overall image count~\cite{chan2008privacy,chan2009bayesian,idrees2013multi} or directly regress the local features to the point coordinates~\cite{song2021rethinking,liang2022end}. Most previous efforts focus on learning a density map, which is a single-channel output with reduced spatial size. It represents the fractional number of objects at each location, and its spatial integration equals the total count of the objects in the image. The density map can be learned by using a pseudo density map generated with Gaussian kernels~\cite{li2018csrnet,liu2019crowd,wan2019adaptive} or directly using a ground-truth point map~\cite{ma2019bayesian,wang2020DMCount,sun2021boosting}.

For architectural choices, 
the past efforts on DOC can also be divided into CNN-based~\cite{li2018csrnet,ranjan2018iterative,liu2019context,liu2019exploiting,laradji2018blobs,song2021rethinking} and Transformer-based methods~\cite{liang2022transcrowd,sun2021boosting,liang2022end}. By nature, convolutional neural networks (CNNs) have limited receptive fields and only use local information. By contrast, Transformers can establish long-range/global relationships among features, whose advantage for DOC has been demonstrated by~\cite{sun2021boosting,liang2022transcrowd,qian2022segmentation}.

\subsection{Multimodal Dense Object Counting}
Multimodal dense object counting (MDOC) differs from DOC by incorporating additional modalities beyond RGB images to perform counting task. These modalities usually include depth images~\cite{9601215,Lian_2019_CVPR,meng2025multi,wang2024multi,li2022rgb,10086642} and thermal images~\cite{liu2021cross,zhang2022spatio,meng2025multi,wu2022multimodal,peng2020rgb,liu2023rgb,mu2024visual}, both containing complementary information and helping to densely count target objects. Rather than solely focusing on designing effective modules to count objects for unimodal approaches, MDOC methods also pay attention to build components which effectively merge information from RGB modality and other additional modalities. For example, CSCA~\cite{zhang2022spatio} employs a cross-modal attention block to fully use multimodal complementarities, by first building connections between different modalities and then aggregating complementary information. IADM~\cite{liu2021cross} contains a information aggregation-distribution module to aggregate and distribute information among multiple modalities. Though these multimodal methods show improved performance compared with unimodal counterparts, they are not designed for IOC task and do not work well when counting objects under challenging camouflaged scenes.

\subsection{Indiscernible Object Counting}
% \vspace{-3pt}
Recently, indiscernible/camouflaged scene understanding has become a popular research topic~\cite{fan2021concealed,le2021camouflaged,le2019anabranch,lamdouar2020betrayed,zhong2022detecting,le2021camouflaged}. It contains a set of tasks specifically focusing on detection, instance segmentation and video object detection/segmentation. It aims to analyze scenes with objects that are difficult to recognize visually~\cite{fan2020Camouflage,lamdouar2020betrayed}. 

In this paper, we study the new task of indiscernible object counting (IOC), which lies at the intersection of dense object counting (DOC) and indiscernible scene understanding. Recently proposed datasets~\cite{yunqiu_cod21,le2021camouflaged,fan2021concealed} for concealed scene understanding can be used as benchmarks for IOC by converting instance-level masks to points. However, they have several limitations, as discussed in~\secref{sec:intro}. Therefore, we propose {the first} large-scale dataset for IOC, \ourdataset.

The ability to estimate depth is nature for human and animals. Their vision systems have effectively evolved to use both RGB and depth signals for understanding the environments. To perform accurate counting in challenging camouflaged scenes, depth maps could largely help by providing rich and complementary information. Therefore, in this paper, we study IOC by additionally using depth modality. To facilitate the research, we introduce the first large-scale RGB-D dataset for IOC, \ourdatasetDepth.
%-------------------------------------------------------------------------

\section{The \ourdataset~Dataset}\label{sec:camfish5k_dataset}
% \vspace{-3pt}
\subsection{Image Collection}\label{sec:img_coll}
% \vspace{-3pt}
Underwater scenes contain many indiscernible objects (\textit{Sea Horse}, \textit{Reef Stonefish}, \textit{Lionfish}, and \textit{Leafy Sea Dragon}) because of limited visibility and active mimicry. Hence, we focus on collecting images of underwater scenes. 

We started by collecting Youtube videos of underwater scenes, using general keywords (\textit{underwater scene}, \textit{sea diving}, \textit{deep sea scene}, \etc.) and category-specific ones (\textit{Cuttlefish}, \textit{Mimic Octopus}, \textit{Anglerfish},  \textit{Stonefish}, \etc.). In total, we collected 135 high-quality videos with lengths from tens of seconds to several hours. Next, we kept one image in every 100 frames ($3.3$ sec) to avoid duplicates. This still leaded to a large number of images, some showing similar scenes or having low quality. Hence, at the final step of image collection, 6 professional annotators carefully reviewed the dataset and removed those unsatisfactory images. The final dataset has 5,637 images, some of which are shown in Fig.~\ref{fig:dataset_samples}. This step cost a total of 200 human hours.

\begin{figure*}[!t]
    \centering
    \includegraphics[width=.999\linewidth]{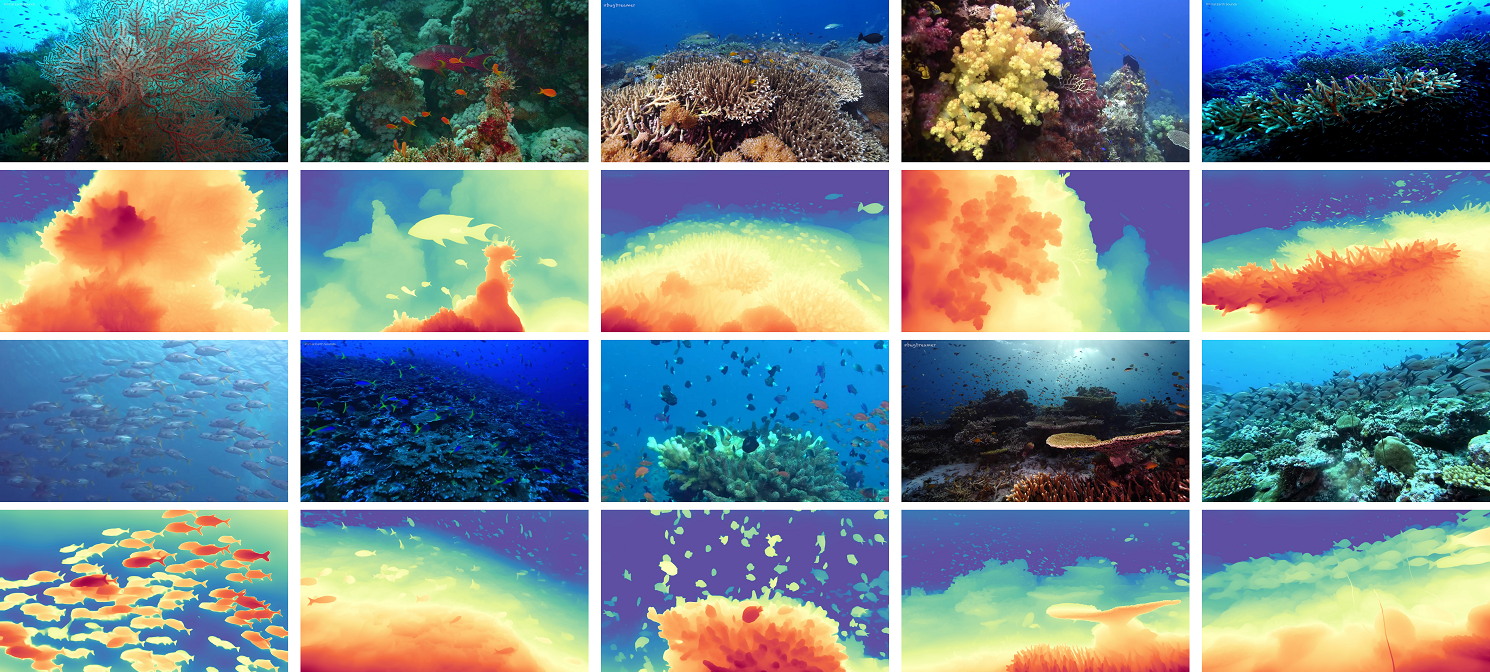}
     \caption{\textbf{Example images and corresponding depth maps from the proposed \ourdatasetDepth.} Each depth map is of high quality and contains detailed distance information about the scene. \textit{Best viewed with zooming.}
     }
    \label{fig:dataset_samples_depth}
\end{figure*}

\subsection{Image Annotation}\label{sec:img_anno}
% \vspace{-3pt}
\myPara{Annotation principles.} The goal was to annotate each animal with a point at the center of its visible part. We have striven for \textit{accuracy} and \textit{completeness}. The former indicates that the annotation point should be placed at the object center, and each point corresponds to exactly one object instance. The latter means that no objects should be left without annotation.

\myPara{Annotation tools.}
To ease annotation, we developed a tool based on open-source Labelimg\footnote{\url{https://github.com/heartexlabs/labelImg}}. It offers the following functions: generate a point annotation in an image by clicking, drag/delete the point, mark the point when encountering difficult cases, and zoom in/out. These functions help annotators to produce high-quality point annotations and to resolve ambiguities by discussing the marked cases.

\myPara{Annotation process.} The whole process is split into \textit{three} steps. First, all annotators (6 experts) were trained to familiarize themselves with their tasks. They were instructed about sea animals and well-annotated samples. Then each of them was asked to annotate 50 images. The annotations were checked and evaluated. When an annotator passed the evaluation, he/she could move to the next step. Second, images were distributed to 6 annotators, giving each annotator responsibility over part of the dataset. The annotators were {required} to discuss {confusing} cases and reach a consensus. Last, they checked and refined the annotations in two rounds. The second step cost 600 human hours, while each checking round in the third step cost 300 hours. The total cost of annotation process amounted to 1,200 human hours.

\begin{table}[t!]
\centering
\caption{\textbf{Comparison of datasets \textit{w.r.t.} image distribution across various density (count) ranges.} We compute the number of images for each dataset under four density ranges.}
\label{table:statis_density}
\small
%\resizebox{\linewidth}{!}{%
\renewcommand{\arraystretch}{1.35}
\renewcommand{\tabcolsep}{1.5mm}
\begin{tabular}{l|c|c|c|c|c}
\hline
\multicolumn{1}{l|}{Datasets} & \multicolumn{1}{c|}{\begin{tabular}[c]{@{}c@{}}\# IMG\\ (0-50)\end{tabular}} & \multicolumn{1}{c|}{\begin{tabular}[c]{@{}c@{}}\# IMG\\ (51-100)\end{tabular}} & \multicolumn{1}{c|}{\begin{tabular}[c]{@{}c@{}}\# IMG\\ (101-200)\end{tabular}} & \multicolumn{1}{c|}{\begin{tabular}[c]{@{}c@{}}\# IMG\\ ($>$200)\end{tabular}} & \multicolumn{1}{c}{Total} \\ \hline
NC4K~\cite{yunqiu_cod21}   &  4,121  &  0   &    0  & 0    &     4,121     \\ 
COD~\cite{fan2021concealed}    & 5,066   &  0   &    0  & 0 &    5,066      \\ 
\textbf{\ourdataset }    &  2,663    &  1,000   &   957   &  1,017   &  5,637         \\ \hline
\end{tabular}
%}

\end{table}

\subsection{Dataset Details}\label{sec:data_statistics}
\begin{figure}[!t]
    \centering
    \includegraphics[width=0.93\linewidth]{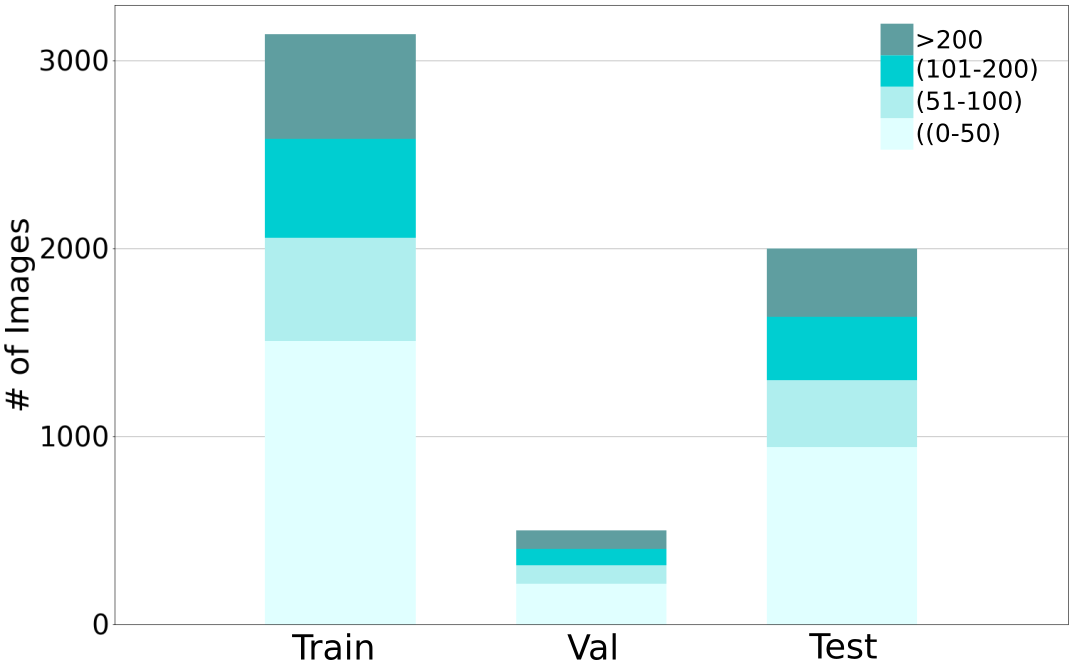}
    \caption{Image distributions under different density (count) ranges ($<$50, 51 to 100, 101 to 200, and $>$200) in training, validation (val), and test sets of \ourdataset.}
    \label{fig:statis_train_val_test}
\end{figure}

% \vspace{-3pt}
The proposed \ourdataset~dataset contains 5,637 high-quality images, annotated with 659,024 points. Table~\ref{table:statis_density} shows the number of images within each count range (0-50, 51-100, 101-200, and above 200). Of all images in \ourdataset, 957 have a medium to high object density, \ie, between 101 and 200 instances. Furthermore, 1,017 images (18\% of the dataset) show very dense scenes ($>200$ objects per image). 
To standardize the benchmarking on \ourdataset, we randomly divide it into three non-overlapping parts: train (3,137), validation (500), and test (2,000).
For each split, the distribution of images across different count ranges follows a similar distribution, as shown in \figref{fig:statis_train_val_test}.

\begin{figure*}[t!]
    \centering
    \includegraphics[width=0.94\linewidth]{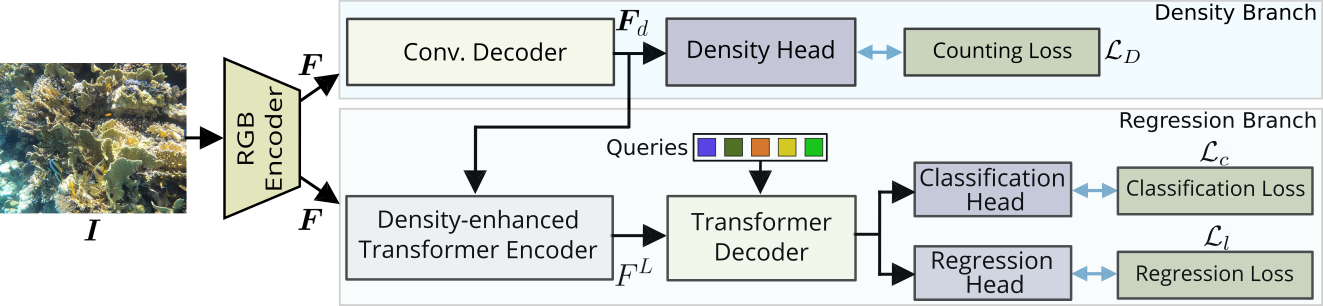}
    % \vspace{-5pt}
     \caption{\textbf{Overview of the proposed \ourmodel.} Given an input image, we extract a feature map using an RGB encoder, which is processed by a density branch and regression branch. The density-enhanced transformer encoder exploits the object density information from the density branch to generate more relevant features for the regression. Refer to \secref{sec:Baseline} for more details.}
     % \vspace{-6pt}
    \label{fig:method}
\end{figure*}

Table~\ref{table:statis} compares the statistics of \ourdataset~with previous datasets. The advantages of \ourdataset~over existing datasets are four-fold. \textbf{(1)} \ourdataset~is the largest-scale object counting dataset for indiscernible scenes. It is superior to its counterparts such as NC4K~\cite{yunqiu_cod21}, CAMO++~\cite{le2021camouflaged}, and COD~\cite{fan2021concealed} in terms of size, image resolution and the number of annotated points. For example, the largest existing IOC dataset CAMO++~\cite{le2021camouflaged} contains a total of 32,756 objects, compared to 659,024 points in \ourdataset. \textbf{(2)} \ourdataset~has far denser images, which makes it currently the most challenging benchmark for IOC. As shown in Table~\ref{table:statis_density}, 1,974 images have more than 100 objects. \textbf{(3)} Although \ourdataset~is specifically proposed for IOC, it has some advantages over the existing DOC datasets. For instance, compared with JHU-CROWD++~\cite{sindagi2020jhu}, which is one of the largest-scale DOC benchmarks, the proposed dataset contains more images with a higher resolution. \textbf{(4)}  \ourdataset~focuses on underwater scenes with sea animal annotations, which makes it different from all existing datasets shown in Table~\ref{table:statis}. Hence, the proposed dataset is {also valuable} for \textit{transfer learning} and \textit{domain adaptation} of DOC~\cite{liu2022leveraging,gong2022bi,chen2021variational,he2021error}.
%  ( and NWPU-Crowd~\cite{gao2020nwpu})

\section{The \ourdatasetDepth~ Dataset}\label{sec:camfish5k_depth}
Recently, vision foundation models have attracted lots of research attention. Different foundation models could have different purposes: some are for segmentation~\cite{kirillov2023segment} while others are for depth estimation~\cite{yang2024depth,yang2024depth2}.
Since the images of \ourdataset~are not captured with depth camera, we consider employing a vision foundation model on depth estimation to generate high-quality pseudo depth maps for \ourdataset. Our dataset with depth images is named \ourdatasetDepth.

\subsection{Pseudo Depth Generation}
\myPara{Depth Anything vs. Depth Anything V2.} We first tried both Depth Anything~\cite{yang2024depth} and Depth Anything V2~\cite{yang2024depth2} models to generate pseudo depth maps for \ourdataset. However, we found that Depth Anything V2 provided significantly better depth maps than its first version. Therefore, we chose to use Depth Anything V2 for examination of depth maps and construction of \ourdatasetDepth, which is explained below.

\myPara{Multi-Scale depth fusion.} There are three different versions of Depth Anything V2 model available, including \textit{small}, \textit{base}, and \textit{large}. We exploited the large model since it outperforms others. For the inference of Depth Anything V2, the input image ($H \times W$) is resized according to a scale factor $s$, given by:
\begin{align}
\begin{split}
    s=\frac{l}{min[H,W]},
\end{split}
\end{align}
where $l$ is a hyperparameter determining the input size used by the model. When using a large $l$, a high-resolution input is forwarded to the network. The default $l$ is 518. We generated the pseudo depth maps on \ourdataset~using different $l$, spanning from 256 to 2048 with a step of 256, and had the following observations after carefully examining the results for hundreds of image examples. \textbf{(1)} The depth maps from large input ($l=1024, 1280$) contain more accurate depth values for small and distant objects since large input contains more details. \textbf{(2)} The depth maps from small input ($l=512, 768$) have more accurate depth values and better boundaries for large and close objects. \textbf{(3)} Too small or large input gives unsatisfactory depth maps probably due to the fact that the model is not trained on such resolutions. Based on these important findings, we designed a multi-scale depth fusion strategy to generate high-quality depth maps for \ourdataset. 

To be more specific, we first generated depth maps using Depth Anything V2 model with four different $l$, including 512, 768, 1024, and 1280. Then the final depth maps from different scales of input images are averaged and normalized to make depth value locate within the range of 0 and 1. The combination of \ourdataset~with final depth maps leads to multimodal \ourdatasetDepth. Some examples of images and corresponding depth maps are shown in \figref{fig:dataset_samples_depth}.

\myPara{Estimated cost.} To examine depth maps and construct \ourdatasetDepth, a total cost of 300 human hours was devoted. In addition, 50 GPU hours were spent to run Depth Anything V2 model. Our dataset \ourdatasetDepth~contains high-quality depth maps, which would facilitate the development of multimodal learning in this field.

% \subsection{Comparison with Other Datasets}
\section{\ourmodel}\label{sec:Baseline}
% \vspace{-3pt}
We first introduce the network structure of our proposed \ourmodel~model, which consists of a density and a regression branch. Then, the novel density-enhanced transformer encoder, which is designed to help the network better recognize and detect indiscernible objects, is explained. 

\subsection{Network Structure}
% \vspace{-3pt}
As mentioned, mainstream methods for object counting fall into two groups: counting-by-density~\cite{wang2020DMCount,li2018csrnet} or counting-by-regression~\cite{liang2022end,song2021rethinking}. The density-based approaches~\cite{wang2020DMCount,li2018csrnet} learn a map with the estimated object density across the image. Differently, the regression-based methods~\cite{liang2022end,song2021rethinking} directly regress to coordinates of object center points, which is straightforward and elegant. As for IOC, foreground objects are difficult to distinguish from the background due to their similar appearance, mainly in color and texture. The ability of density-based approaches to estimate the object density level could be exploited to make (indiscernible) foreground objects stand out and improve the performance of regression-based methods. In other words, the advantages of density-based and regression-based approaches could be combined. Thus, we propose \ourmodel, which contains two branches: a density branch and a regression branch, as in Fig.~\ref{fig:method}. The density branch's information helps refine the regression branch's features.

Formally, we are given an input image $\bm{I}$ with ground-truth object points $\{(x_i,y_i)\}_{i=1}^{T}$ where $(x_i,y_i)$ denotes the coordinates of the $i$-th object point and $T$ is the total number of objects. The goal is to train an object counting model which predicts the number of objects in the image. We first extract a feature map $\bm{F} \in \mathbb{R}^{h \times w \times c_1}$ ($h$, $w$, and $c_1$ denote height, weight, and the number of channels, respectively) by sending the image through an RGB encoder. Next, $\bm{F}$ is processed by the density and the regression branches. 

The density branch inputs $\bm{F}$ into a convolutional decoder which consists of two convolutions with $3\times 3$ kernels. A density-aware feature map $\bm{F}_d \in \mathbb{R}^{h \times w \times c_2}$ is obtained, where $c_2$ is the number of channels. Then a density head (a convolution layer with $1\times 1$ kernel and $\rm ReLU$ activation) maps $\bm{F}_d$ to a single-channel density map $\bm{D} \in \mathbb{R}^{h \times w}$ with non-negative values. Similar to \cite{wang2020DMCount}, the counting loss ($L_1$ loss) used in the density branch is defined as:
\begin{align}\label{Eq:density_loss}
% \footnotesize
% \vspace{-2pt}
\begin{split}
    &\mathcal{L}_D=\big|\lVert \bm{D} \rVert_1 - T\big|,
\end{split}
% \vspace{-2pt}
\end{align}
where $\lVert \cdot \lVert_1$ denotes the entry-wise $L_1$ norm of a matrix. The density map $\bm{D}$ estimates the object density level across the spatial dimensions. Hence, the feature map $\bm{F}_d$ before the density head is density-aware and contains object density information, which could be exploited to strengthen the feature regions with indiscernible object instances.

As to the regression branch, the feature map $\bm{F}$ from the encoder and the density-aware feature map $\bm{F}_d$ from the density branch are first fed into our density-enhanced transformer encoder, described in detail in~\secref{sec:density-enhance-trans-encoder}. After this module, the refined features, together with object queries, are passed to a typical transformer decoder~\cite{vaswani2017attention}. The decoded query embeddings are then used by the classification head and regression head to generate predictions. The details are explained in~\secref{sec:distance-based-classification-loss}.

\begin{figure}
\centering
\subfloat[]{\label{fig:density_trans_encoder_a}\includegraphics[width=0.97\linewidth]{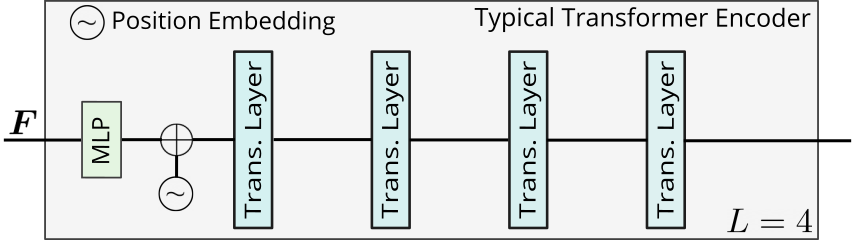}} \\
\subfloat[]{\label{fig:density_trans_encoder_b}\includegraphics[width=0.97\linewidth]{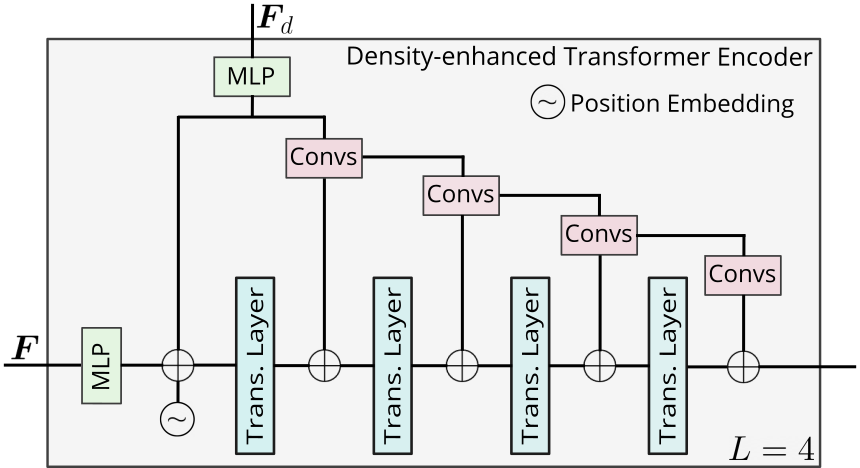}}
%
% \vspace{-7pt}
\caption{\textbf{Comparison between typical transformer encoder (a) and our density-enhanced transformer encoder (b) when $L=4$.}}
% \vspace{-7pt}
\label{fig:density_trans_encoder}
\end{figure}

\begin{figure*}[t!]
    \centering
    \includegraphics[width=0.94\linewidth]{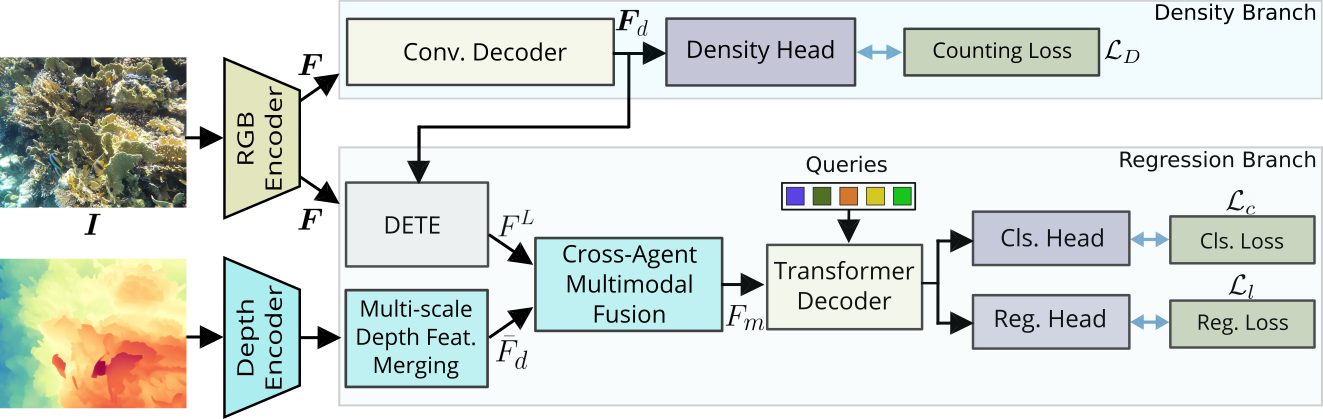}
     \caption{\textbf{Overview of the proposed \ourmodelDepth.} Given an input image and corresponding depth map, the RGB information is processed in a similar way as \ourmodel~while the depth modality is input to a small depth encoder and Multi-scale Depth Feature Merging (MDFM) module. Both RGB and depth features are fused through the proposed Cross-Agent Multimodal Fusion (CAMF) module. Refer to \secref{sec:iocformer-d} for more details.}
    \label{fig:method2}
\end{figure*}

\subsection{Density-Enhanced Transformer Encoder} \label{sec:density-enhance-trans-encoder}
Here, we explain the density-enhanced transformer encoder (DETE) in detail. The structure of the typical transformer encoder (TTE) and the proposed DETE is shown in Fig.~\ref{fig:density_trans_encoder}. Different from TTE, which directly processes one input, DETE takes two inputs: the features ($\bm{F}$) extracted by the initial encoder and the density-aware features ($\bm{F}_d$) from the density branch. DETE uses the density-aware feature map to refine the encoder feature map. With information about which image areas have densely distributed objects and which have sparsely distributed objects, the regression branch can more accurately predict the positions of indiscernible object instances.

We first project $\bm{F}$ to $\hat{\bm{F}} \in \mathbb{R}^{h \times w \times c}$, and $\bm{F}_d$ to $\hat{\bm{F}}_d \in \mathbb{R}^{h \times w \times c}$ by using an MLP layer so that the number of channels ($c$) matches. The input to the first transformer layer is the combination of $\hat{\bm{F}}$, $\hat{\bm{F}}_d$ and position embedding $E \in \mathbb{R}^{hw\times c}$. This process is given by:
\begin{align}
\begin{split}
        &{F}^1={\rm Rs}(\hat{\bm{F}})+{\rm Rs}(\hat{\bm{F}}_d)+E;~~F^2={\rm Trans}(F^1),\\
\end{split}
\end{align}
where ${\rm Rs}(\cdot)$ denotes the operation of reshaping the feature map by flattening its spatial dimensions, and ${\rm Trans}(\cdot)$ denotes a transformer layer. 
After that, additional transformer layers are used to further refine the features, as follows:

\begin{align}\label{Eq:tranformers}
\begin{split}
        % &F^1=\hat{F}+\hat{\bm{F}}_d+E,\\
        &\bm{F}_d^{1}=\hat{\bm{F}}_d,\\
        &\bm{F}_d^{i}={\rm Convs}(\bm{F}_d^{i-1}),~~i=2,3,...,L-1,\\
        &F^{i+1}={\rm Trans}(F^i+{\rm Rs}(\bm{F}_d^{i})),~~i=2,3,...,L-1,\\
\end{split}
\end{align}
where $\rm {Convs}(\cdot)$ denotes a convolutional block containing two convolution layers. The total number of transformer layers is $L$, which also represents the total times of merging transformer and convolution features. After \equref{Eq:tranformers}, we obtain the density-refined features $F^L \in \mathbb{R}^{hw \times c}$, which are then forwarded to the transformer decoder.

The benefit of our DETE can also be interpreted from the perspective of \textit{global} and \textit{local} information. Before each transformer layer in \equref{Eq:tranformers}, we merge features from the previous transformer layer (global) and features from the convolutional block (local). 
During this process, the global and local information gradually get combined, which boosts the representation ability of the module.

\subsection{Loss Function}\label{sec:distance-based-classification-loss}
% \vspace{-3pt}
After the DETE module, we obtain density-refined features $F^L$. Next, the transformer decoder takes the refined features $F^L$ and trainable query embeddings $Q \in \mathbb{R}^{n \times c}$ containing $n$ queries as inputs, and outputs decoded embeddings $\hat{Q} \in \mathbb{R}^{n \times c}$. The transformer decoder consists of several layers, each of which contains a self-attention module, a cross-attention layer and a feed-forward network (FFN). For more details, we refer to the seminal work~\cite{vaswani2017attention}. $\hat{Q}$ contains $n$ decoded representations, corresponding to $n$ queries. Following \cite{liang2022end}, every query embedding is mapped to a confidence score by a classification head and a point coordinate by a regression head. Let $\{p_i, (\hat{x}_i, \hat{y}_i)\}_{i=1}^{n}$ denote the predictions for all queries, where $p_i$ is the predicted confidence score determining the likelihood that the point belongs to the foreground and $(\hat{x}_i, \hat{y}_i)$ is the predicted coordinate for the $i$-th query. Then we conduct a Hungarian matching~\cite{carion2020end,liang2022end} between predictions $\{p_i, (\hat{x}_i, \hat{y}_i)\}_{i=1}^{n}$ and ground-truth $\{(x_i,y_i)\}_{i=1}^{T}$. Note that $n$ is bigger than $T$ so that each ground-truth point has a matched prediction. The Hungarian matching is based on the $k$-nearest-neighbors matching objective~\cite{liang2022end}. Specifically, the matching cost depends on three parts: the distance between predicted points and ground-truth points, the confidence score of the predicted points, and the difference between predicted and ground-truth average neighbor distance~\cite{liang2022end}. After the matching, we compute the classification loss $\mathcal{L}_c$, which boosts the confidence score of the matched predictions and suppresses the confidence score of the unmatched ones. To supervise the predicted coordinates' learning, we also compute the localization loss $\mathcal{L}_l$, which measures the $L_1$ distance between the matched predicted coordinates and the corresponding ground-truth coordinates. For more details, we refer to \cite{liang2022end}. The final loss function is defined as:
\begin{align}\label{Eq:loss}
\begin{split}
        &\mathcal{L}=\lambda \mathcal{L}_D+\mathcal{L}_c+\mathcal{L}_l,
\end{split}
\end{align}
where $\lambda$ is set to 0.5. The density and the regression branch are jointly trained using \equref{Eq:loss}. During inference, we take the predictions from the regression branch.

\section{\ourmodelDepth}\label{sec:iocformer-d}
Depth estimation is an innate ability of the human binocular vision system. This type of information is inherently used by the human brain to understand real-world environments. For common scenes where the objects are easy to distinguish, the RGB modality can already provide excellent perception performance. However, for challenging camouflaged scenarios where the information contained in RGB images is not sufficient, depth maps can provide valuable complementary information. Therefore, we consider exploiting depth maps for the studied IOC task and propose \ourmodelDepth, which is plotted in \figref{fig:method2}. The details of \ourmodelDepth~are explained in the following.

\subsection{Multi-scale Depth Feature Merging}\label{sec:multi-scale-depth-feature}
Following the notations in \secref{sec:Baseline}, we also have a depth map $\bm{M}$ associated with the input image $\bm{I}$. Besides the RGB encoder to extract feature maps for $\bm{I}$, we also have a small depth encoder, \eg, PVTv2~\cite{wang2022pvt} or Swin~\cite{liu2021swin}, to extract multi-scale depth feature maps $\bm{F}_d=\{\bm{F}_d^i\}_{i=1}^{S}$ from $\bm{M}$, where $S$ is the number of intermediate features. We use a Multi-scale Depth Feature Merging (MDFM) module to aggregate intermediate feature maps in different scales. 

First, each feature map $\bm{F}_d^i$ is processed using a linear layer and an interpolation operation to resize it, so that the output has the same spatial size ($h_d\times w_d$) and same number of channels ($\frac{c}{S}$). This process is denoted as:
\begin{align}\label{Eq:mdfm}
\begin{split}
   &\hat{\bm{F}}_d^i={\rm Resize}({\rm Linear}(\bm{F}_d^i)),~i \in [1, S],
\end{split}
\end{align}
where each $\hat{\bm{F}}_d^i \in \mathbb{R}^{h_d\times w_d\times \frac{c}{S}}$.
After that, we concatenate all feature maps into a single feature map which contains comprehensive information from different intermediate layers. Next, we resize the obtained feature to the spatial resolution of $h \times w$, same as $F^L$ (feature output from DETE). The mentioned operations are as follows:
\begin{align}\label{Eq:mdfm_merge}
\begin{split}
   &\bar{\bm{F}}_d={\rm Resize}({\rm Cat}([\hat{\bm{F}}_d^1, ..., \hat{\bm{F}}_d^i, ..., \hat{\bm{F}}_d^S])),
\end{split}
\end{align}
where {\rm Cat()} represents feature concatenation along the channel dimension, and $\bar{\bm{F}}_d \in \mathbb{R}^{h\times w\times c}$. The generated feature map $\bar{\bm{F}}_d$ originates from the depth modality $\bm{M}$, containing rich and complementary information about the scene. To prepare for the attention operation, we reshape $\bar{\bm{F}}_d$ to the size of $\mathbb{R}^{hw\times c}$, denoted by $\bar{{F}}_d$. In the following, we will introduce how to fuse information from RGB and depth modality.

\begin{figure}
\centering
\subfloat[]{\label{fig:cross_modal_fusion_a}\includegraphics[width=0.97\linewidth]{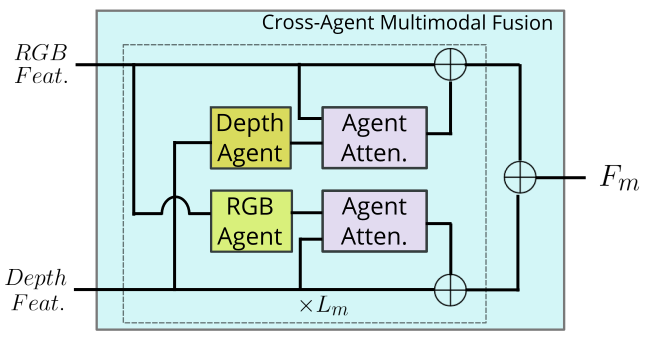}} \\
\subfloat[]{\label{fig:cross_modal_fusion_b}\includegraphics[width=0.97\linewidth]{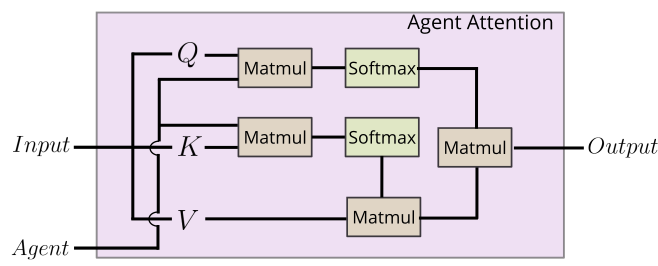}}
\caption{\textbf{Illustration of the proposed Cross-Agent Multimodal Fusion module (a) and Agent Attention (b).}}
\label{fig:cross_modal_fusion}
\end{figure}

\subsection{Cross-Agent Multimodal Fusion}\label{sec:cross-agent-multimodal-fusion}
We have the RGB feature $F^L$ from the input image $\bm{I}$ and the depth feature $\bar{{F}}_d$ from the associated depth map $\bm{M}$. The goal is to effectively fuse both features from different modalities. Therefore, we propose a Cross-Agent Multimodal Fusion (CAMF) module, the structure of which is shown in \figref{fig:cross_modal_fusion}. Our CAMF is inspired by agent attention~\cite{han2025agent}. The agent attention improves typical self-attention by introducing agents to perform information exchange between the input tokens, achieving favorable balance between computations and representation capability~\cite{han2025agent}. In the following, we first briefly introduce agent attention and then explain the proposed CAMF for effectively merging multimodal information.

\myPara{Agent attention.} For typical self-attention adopted in Transformers~\cite{vaswani2017attention}, affinity map is computed among all query and key tokens, leading to large time complexity of $O(N^2)$. To improve this, agent attention~\cite{han2025agent} introduces agent tokens to model the relationship among the input, which has the time complexity of $O(NN_a)$. Here, $N_a$ is the number of agent tokens and $N_a < N$. The operations of agent attention are shown in \figref{fig:cross_modal_fusion_b}.

Formally, given \textit{Input} tokens and \textit{Agent} tokens ($A$), agent attention models the complex relationships among them and generate the refined tokens ($Output$). Let's denote query $Q \in \mathbb{R}^{N\times c}$, key $K \in \mathbb{R}^{N\times c}$, value $V \in \mathbb{R}^{N\times c}$, and agent $A \in \mathbb{R}^{N_a\times c}$. The output $O \in \mathbb{R}^{N\times c}$ is given by:
\begin{align}\label{Eq:agentattn}
\begin{split}
   &O= {\rm AAttn}(Input,~A)=\sigma(QA^T)\sigma(AK^T)V,
\end{split}
\end{align}
where AAttn represents agent attention and $\sigma$ represents softmax function along the last dimension. When taking feature dimension $c$ into consideration, \equref{Eq:agentattn} has the time complexity of $O(NN_{a}c)$, which is less than $O(N^2c)$ of self-attention. According to \cite{han2025agent}, agent tokens ($A$) can be generated from the input tokens through simple pooling strategy. We analyze that the agent $A$ contains condensed information about the input, which can serve as an intermediary for information exchange between RGB and depth modalities. Therefore, we propose cross-agent multimodal fusion module to aggreate RGB and depth features through the agent attention.

\myPara{CAMF module.} CAMF integrates the RGB feature ($F^L$) and depth feature ($\bar{{F}}_d$) through cross-agent attention. The illustration of this module is depicted in \figref{fig:cross_modal_fusion_a}. For the sake of easy explanation, we re-denote the RGB and depth features (inputs to the CAMF) as $F^0_{rgb}$ and $F^0_{depth}$, respectively. 

First, RGB and depth features are first used to generate RGB and depth agents, $A^0_{rgb}$ and $A^0_{depth}$, respectively. Next, the RGB feature and $A^0_{depth}$ are passed to a agent attention to generate refined RGB feature. Similarly, the depth feature and $A^0_{rgb}$ are passed to a agent attention to generate refined depth feature. This process is given by:
\begin{align}
\begin{split}
   &F^1_{rgb}= {\rm AAttn}(F^0_{rgb},~A^0_{depth}),\\
   &F^1_{depth}= {\rm AAttn}(F^0_{depth},~A^0_{rgb}).
\end{split}
\end{align}
The above step is repeated for $L_m$ times for extensive information exchange between different modalities. We obtain the final refined RGB and depth features $F^{L_m}_{rgb}$ and $F^{L_m}_{depth}$. Next, two features are merged into $F_{m}$ by simple addition, as follows:
\begin{align}
\begin{split}
   &F_m= F^{L_m}_{rgb}+F^{L_m}_{depth}.
\end{split}
\end{align}
The merged feature $F_m$ contains complementary information from different modalities, which is propagated to the Transformer decoder, followed by loss computation, as discussed in \S\ref{sec:distance-based-classification-loss}. 

\begin{table*}[!t]
\centering
\caption{\textbf{Comparison with unimodal state-of-the-art methods on the validation and test set of \ourdataset.} \ourmodel~clearly outperforms exsiting approaches. The best results are highlighted in \textbf{bold}.}
\label{table:results_baselines}
% \footnotesize
\small
\renewcommand{\arraystretch}{1.35}
\renewcommand{\tabcolsep}{4.2mm}
%\setlength{\tabcolsep}{12pt}
% \resizebox{1.0\textwidth}{!}{%
\begin{tabular}{l|c|c|c|c|c|c|c}
\hline
% \toprule
% \rowcolor{mygray}
 & & \multicolumn{3}{c|}{Val (500)}                          & \multicolumn{3}{c}{Test (2,000)}      \\ 
 % \cmidrule{3-8} 
 \hhline{*{2}{|~}*{6}{|-}}
% \hhline{|*2{>{\arrayrulecolor{mygray}}-}>{\arrayrulecolor{black}}|*4{-}|}
% \rowcolor{mygray}
{\multirow{-2}{*}{Method}}  & {\multirow{-2}{*}{Publication}}  & \multicolumn{1}{c|}{~~MAE$\downarrow$~~} & \multicolumn{1}{c|}{~~MSE$\downarrow$~~} & \multicolumn{1}{c|}{~~NAE$\downarrow$~~} & \multicolumn{1}{c|}{~~MAE$\downarrow$~~} & \multicolumn{1}{c|}{~~MSE$\downarrow$~~} & \multicolumn{1}{c}{~~NAE$\downarrow$~~} \\ \hline 
% mandate
MCNN~\cite{zhang2016single} & CVPR'16 & 81.62 & 152.09 &3.53   &72.93   &129.43 &  4.90\\ 
%\hline
CSRNet~\cite{li2018csrnet} & CVPR'18 & 43.05 & 78.46 &1.91   &38.12   &69.75 &  2.48\\ 
%\hline
LCFCN~\cite{laradji2018blobs} & ECCV'18 & 31.99 & 81.12 &0.77   &28.05   &68.24 &  1.12\\ 
%\hline
CAN~\cite{liu2019context} & CVPR'19 & 47.77 & 83.67 &2.10   &42.02   &74.46 &  2.58\\ 
%\hline
DSSI-Net~\cite{liu2019crowd} & ICCV'19 & 33.77 & 80.08 &1.25   &31.04   &69.11 &  1.68\\ 
%\hline

% mandate
BL~\cite{ma2019bayesian} & ICCV'19 & 19.67 & 44.21 &0.39   &20.03   &46.08 &  0.55\\ 
%\hline
NoisyCC~\cite{wan2020modeling} & NeurIPS'20 & 19.48 & 41.76 &0.39   &19.73   &46.85 &  0.46\\ 
%\hline
DM-Count~\cite{wang2020DMCount}   & NeurIPS'20 & 19.65 & 42.56 &0.42   &19.52   &45.52 &  0.55\\ 
%\hline
GL~\cite{wan2021generalized} & CVPR'21 & 18.13 & 44.57 & 0.33  & 18.80  & 46.19 & 0.47 \\
% 18.13333333	44.57	0.33	18.8	46.19	0.4733333333
P2PNet~\cite{song2021rethinking} & ICCV'21 & 21.38 & 45.12 &0.39   &20.74   &47.90 &  0.48\\ 
%\hline
KDMG~\cite{9189836} & TPAMI'22 & 22.79 & 47.32 & 0.90  & 22.79  & 49.94 & 1.17 \\ 
MPS~\cite{zand2022multiscale} & ICASSP'22 & 34.68 & 59.46 &2.06   &33.55   &55.02 &  2.61\\ 
%\hline
MAN~\cite{lin2022boosting} & CVPR'22 & 24.36 & 40.65 &2.39   &25.82   &45.82 &  3.16\\ 
%\hline
CLTR~\cite{liang2022end} & ECCV'22 & 17.47 & 37.06 & 0.29  & 18.07  & 41.90 &  0.43 \\ \hline 
%\hline %
\textbf{\ourmodel~(Ours)} & CVPR'23 & \textbf{15.91} & \textbf{34.08} & \textbf{0.26} & \textbf{17.12}  & \textbf{41.25}  & \textbf{0.38} \\ \hline
\end{tabular}
% }
\end{table*}

\section{Experiments}\label{sec:experiments}
% \vspace{-3pt}
\subsection{Experimental Setting}
% \vspace{-3pt}
\myPara{Compared unimodal methods.} Since there is no algorithm specifically designed for IOC, we select 14 recent open-source DOC methods for benchmarking. Selected methods include: MCNN~\cite{zhang2016single}, CSRNet~\cite{li2018csrnet}, LCFCN~\cite{laradji2018blobs}, CAN~\cite{liu2019context}, DSSI-Net~\cite{liu2019crowd}, BL~\cite{ma2019bayesian}, NoisyCC~\cite{wan2020modeling}, DM-Count~\cite{wang2020DMCount}, GL~\cite{wan2021generalized}, P2PNet~\cite{song2021rethinking}, KDMG~\cite{9189836}, MPS~\cite{zand2022multiscale}, MAN~\cite{lin2022boosting}, and CLTR~\cite{liang2022end}. 
Among them, P2PNet and CLTR are based on regression, while others are on density map estimation. Their codes are all publicly available. The details of these methods are as follows:
\begin{itemize}
    \item MCNN: It proposes a multi-column convolutional neural network that contains different convolution branches with different receptive fields. The ground-truth density map is calculated using geometry-adaptive kernels. 
    \item {CSRNet:} It aims at conducting crowd counting under highly congested scenes. CSRNet exploits dilated convolutions in this task and achieve promising results.
    \item {LCFCN:} This method predicts a blob for each object instance by using only point supervision. It achieves excellent performance in crowd counting as well as generic object counting.
    \item {CAN:} CAN processes encoded features (VGG-16) with different receptive fields, which are then combined using the learned weights. The final context-aware features are passed to estimate the density map. 
    \item {DSSI-Net:} It focuses on tackling the problem of large-scale variation in crowd counting and proposes structured feature enhancement and dilated multi-scale structural similarity loss to generate better density maps.
    \item {BL:} Different from previous works which adopt $L_1$ or $L_2$ loss for supervising the learning of density maps, BL proposes a Bayesian loss which directly uses point annotations to learn density probability.
    \item {NoisyCC:} NoisyCC explicitly models the annotation noise in crowd counting with a joint Gaussian distribution. A low-rank covariance approximation is derived to improve the efficiency~\cite{wan2020modeling}.
    \item {DM-Count:} This method proposes to exploit distribution matching for crowd counting. The optimal transport algorithm is used to minimize the gap between the predicted density map and the ground-truth point map.
    \item {GL:} GL proposes a perspective-guided optimal transport cost function for crowd counting. It is currently the most powerful loss for crowd counting and achieves state-of-the-art performance on mainstream DOC datasets compared to other loss functions.
    \item {P2PNet:} It directly predicts a number of point proposals (location and confidence score). Then Hungarian algorithm~\cite{carion2020end} is used to match proposals and point annotations. It is a purely point-based algorithm for crowd counting~\cite{song2021rethinking} and achieves impressive performance on DOC datasets. 
    \item {KDMG:} Different from previous density-based methods, which generates ground-truth density map by convolving the point map with a/an (adaptive) Gaussian kernel, KDMG proposes a density map generator that is jointly trained with counting model. 
    \item {MPS:} This method generates multi-scale features for the crowd image and benefits from the joint learning of crowd counting as well as localization.
    \item {MAN:} It deals with the problem of large-scale variations in crowd counting by integrating global attention, local attention, and instance attention in a unified framework. MAN achieves state-of-the-art performance on mainstream datasets such as JHU++ and NWPU.
    \item {CLTR:} It directly predicts the point locations by adopting a transformer encoder and decoder structure to process the features. The trainable embeddings are used to extract object locations from the encoded features.
\end{itemize}

\myPara{Compared multimodal methods.} We use 4 open-source multimodal DOC methods for benchmarking \ourdatasetDepth, including RDNet~\cite{Lian_2019_CVPR}, IADM~\cite{liu2021cross}, CSCA~\cite{zhang2022spatio}, and BM~\cite{meng2025multi}. IADM was originally designed for RGBT counting, and was adapted for our use case. The details of these methods are listed below:
\begin{itemize}
    \item RDNet: It performs object counting and localization simultaneously, by using a regression-guided detection network. Depth information is exploited to generate high-quality density map and initialize anchor sizes for detection.
    \item IADM: This approach focuses on learning cross-modal representations which capture complementary information from different modalities, through an information aggregation-distribution module.
    \item CSCA: It proposes a cross-modal spatio-channel attention block which first models global feature correlations across different modalities and then aggregates complementary features.
    \item BM: To mitigate the gap between different modalities, the recent BM method introduces an virtual auxiliary broker modality and considers the multimodal learning as a triple-modal problem.
\end{itemize}

\begin{table*}[!t]
\centering
\caption{\textbf{Comparison with multimodal state-of-the-art methods on the validation and test set of \ourdatasetDepth.} \ourmodelDepth~clearly outperforms existing approaches.}
\label{table:results_baselines-depth}
% \footnotesize
\small
\renewcommand{\arraystretch}{1.35}
\renewcommand{\tabcolsep}{4.2mm}
%\setlength{\tabcolsep}{12pt}
% \resizebox{1.0\textwidth}{!}{%
\begin{tabular}{l|c|c|c|c|c|c|c}
\hline
% \toprule
% \rowcolor{mygray}
 & & \multicolumn{3}{c|}{Val (500)}                          & \multicolumn{3}{c}{Test (2,000)}      \\ 
 % \cmidrule{3-8} 
 \hhline{*{2}{|~}*{6}{|-}}
% \hhline{|*2{>{\arrayrulecolor{mygray}}-}>{\arrayrulecolor{black}}|*4{-}|}
% \rowcolor{mygray}
{\multirow{-2}{*}{Method}}  & {\multirow{-2}{*}{Publication}}  & \multicolumn{1}{c|}{~~MAE$\downarrow$~~} & \multicolumn{1}{c|}{~~MSE$\downarrow$~~} & \multicolumn{1}{c|}{~~NAE$\downarrow$~~} & \multicolumn{1}{c|}{~~MAE$\downarrow$~~} & \multicolumn{1}{c|}{~~MSE$\downarrow$~~} & \multicolumn{1}{c}{~~NAE$\downarrow$~~} \\ \hline 
% mandate

RDNet~\cite{Lian_2019_CVPR}  & CVPR'19 & 25.79 & 60.85 & 1.29  & 25.27  & 56.69 & 1.43  \\ 
IADM~\cite{liu2021cross}  & CVPR'21 & 20.27 & 41.32 & 0.80  & 20.67  & 44.93 & 0.95  \\ 
CSCA~\cite{zhang2022spatio} & ACCV'22 & 24.42 & 55.04 & 0.98  & 24.1  & 50.97 & 1.24  \\ 
BM~\cite{meng2025multi}  & ECCV'24 & 18.77 & 43.69 & 0.74  & 18.45  & 40.88 & 0.86  \\ \hline 
 \textbf{\ourmodelDepth~(Ours)}  & - & \textbf{15.19} & \textbf{32.89} & \textbf{0.24}  & \textbf{16.80}  & \textbf{40.60} & \textbf{0.33}  \\ \hline 
\end{tabular}
% }
\end{table*}

\myPara{Implementation details of unimodal methods.} 
For methods such as MCNN and CAN, we use open-source re-implementations for our experiments. For the other methods, we use official codes and default parameters.
All experiments are conducted on PyTorch~\cite{paszke2019pytorch} and NVIDIA GPUs. $L$ in DETE is set to 6 and the number of queries ($n$) is set as 700. Following~\cite{liang2022end}, our \ourmodel~uses ResNet-50~\cite{he2016deep} as encoder, pretrained on Imagenet~\cite{deng2009imagenet}. Other modules/parameters are randomly initialized. For data augmentations, we use random resizing and horizontal flipping. The images are randomly cropped to $256 \times 256$ inputs. Each batch contains 8 images, and the Adam optimizer~\cite{kingma2014adam} is used. During inference, we split the images into patches of the same size as during training. Following \cite{liang2022end}, we use a threshold (0.35) to filter out background predictions.

\myPara{Implementation details of multimodal methods.} For the compared multimodal methods, we use official implementations and default parameters. For multimodal \ourmodelDepth, the depth encoder we use is PVTv2-B0~\cite{wang2022pvt}, which is the smallest version of such backbones. The feature maps from the last four stages are used in multi-scale depth feature merging step (\S\ref{sec:multi-scale-depth-feature}). For $L_m$ in cross-agent multimodal fusion module (\S\ref{sec:cross-agent-multimodal-fusion}), we use 2 times of feature interactions, which leads to good performance. For all other hyper-parameters, we follow the settings in \ourmodel~for fair comparisons.

\myPara{Evaluation metrics.} To evaluate the effectiveness of the baselines and the proposed method, we compute Mean Absolute Error (MAE), Mean Square Error (MSE), and Mean Normalized Absolute Error (NAE) between predicted counts and ground-truth counts for all images, following~\cite{wang2020DMCount,liang2022end,gao2020nwpu}.

% Train set: 3137. Total: 5637

\subsection{Unimodal Counting Results and Analysis}
% \vspace{-3pt}
We present the results of 14 mainstream crowd-counting algorithms and \ourmodel~in Table~\ref{table:results_baselines}. All methods follow the same evaluation protocol: the model is selected via the val set.
Based on the results, we observe:
\begin{itemize}
    % \vspace{-5.1pt}
    \item Among all previous methods, the recent CLTR~\cite{liang2022end} outperforms the rest, with 18.07, 41.90, 0.43 on the test set for MAE, MSE, and NAE, respectively. The reason is that this method uses a transformer encoder to learn global information and a transformer decoder to directly predict center points for object instances.
    % \vspace{-6pt}
    \item Some methods (MAN~\cite{lin2022boosting} and P2PNet~\cite{song2021rethinking}) perform competitively on DOC datasets such as JHU++~\cite{sindagi2019pushing} and NWPU~\cite{gao2020nwpu}, but perform worse on \ourdataset. For example, MAN achieves 53.4 and 209.9 for MAE and MSE on JHU++, outperforming other methods, including CLTR which achieves 59.5 and 240.6 for MAE and MSE. However, MAN underperforms on \ourdataset, compared to CLTR, DM-Count, NoisyCC, and BL. This shows that methods designed for DOC do not necessarily work well for indiscernible objects. Hence, IOC requires specifically designed solutions.
    % \vspace{-6pt}
    \item These methods, including BL, NoisyCC, DM-Count, and GL, which propose new loss functions for crowd counting, perform well despite being simple. For example, GL achieves 18.80, 46.19, and 0.47 for MAE, MSE, and NAE on the test set.
    % \vspace{-6.9pt}
\end{itemize}

Different from previous methods, \ourmodel~is specifically designed for IOC with two novelties: (1) combining density and regression branches in a unified framework, which improves the underlying features; (2) density-based transformer encoder, which strengthens the feature regions where objects exist. On both the val and test sets, \ourmodel~is superior to all previous methods for MAE, MSE, and NAE. 

\begin{table}[!t]
\centering
\caption{\textbf{Impact of density branch (DB) and DETE on \ourdataset~val set.} For DB+Regression without using DETE, a typical transformer encoder (TTE) is used instead.}
\label{table:ablation_structure}
% \footnotesize
\small
\renewcommand{\arraystretch}{1.35}
\renewcommand{\tabcolsep}{3.0mm}
%\setlength{\tabcolsep}{8pt}
%\resizebox{0.48\textwidth}{!}{%
\begin{tabular}{l|c|c|c|c}
\hline
Methods     & DETE & MAE$\downarrow$ & MSE$\downarrow$ & NAE$\downarrow$ \\ \hline
DB      &  \xmark    &  18.25   & 39.77   & 0.29 \\ 
Regression             &  \xmark    &  17.47   &  37.06   &  0.29   \\ \hline
\multirow{2}{*}{DB+Regression} &  \xmark    &  16.94   &  35.92   & \textbf{0.26}    \\ %\cline{2-5} 
     &  \cmark    & \textbf{15.91}    & \textbf{34.08}    & \textbf{0.26}    \\ \hline
     %  &  \cmark    &  15.75   &  35.22   &  0.26   \\ \hline
    %  [15.750] mse [35.225] nae [0.2615]
\end{tabular}
%}
\end{table}

\subsection{Multimodal Counting Results and Analysis}
We also present the results of 4 popular multimodal counting algorithms and \ourmodelDepth~in Table~\ref{table:results_baselines-depth}. From the results, we have the following observations:
\begin{itemize}
    \item Among the considered multimodal counting methods, the recent model BM clearly outperforms others (RDNet, IADM and CACA), possibly due to its excellent ability of merging useful information from different modalities through an auxiliary broker modality.
    \item Though multimodal counting methods exploit depth information, they are still inferior to our unimodal method \ourmodel. The reason is that they are not specifically designed for camouflaged scenes. This demonstrates the necessity and importance of studying object counting under camouflaged environments.
    \item Due to the effective usage of depth information through CAMF module, our multimodal \ourmodelDepth~clearly surpasses unnimodal \ourmodel, by reducing 0.72 MAE and 1.19 MSE. What's more, \ourmodelDepth~achieves new state-of-the-art results on the challenging counting dataset \ourdatasetDepth.
\end{itemize}

\begin{figure*}[!t]
    \centering
\includegraphics[width=.99\linewidth]{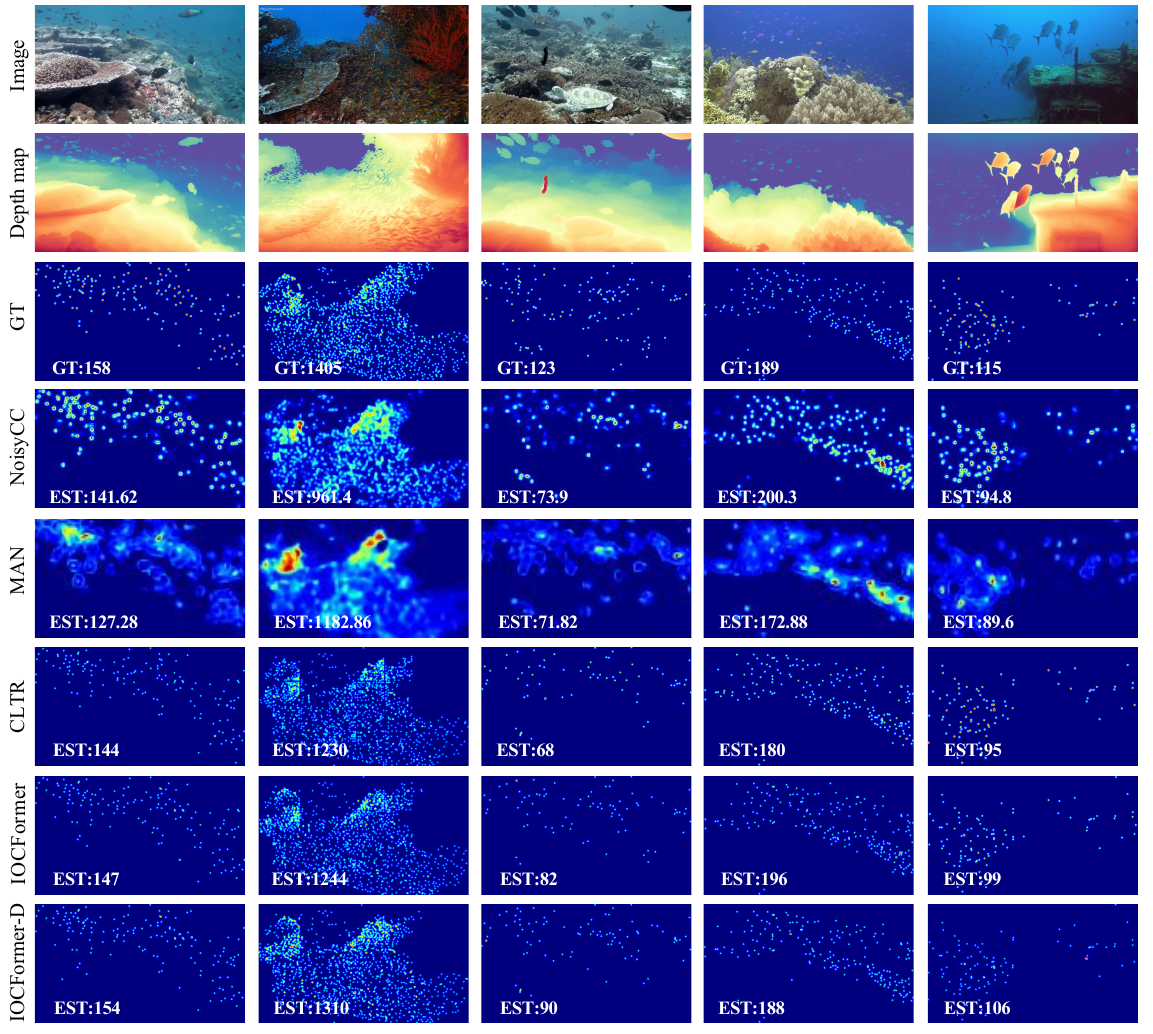}
    % \vspace{-5pt}
     \caption{
     \textbf{Qualitative comparisons of various algorithms: NoisyCC~\cite{wan2020modeling}, MAN~\cite{lin2022boosting}, CLTR~\cite{liang2022end}, \ourmodel~(ours), and \ourmodelDepth~(ours).}
     The GT or estimated counts for each case are shown in the lower left corner. \textit{Best viewed with zooming.}
     }
    \label{fig:qual_examples}
    \vspace{-1mm}
\end{figure*}

Besides the quantitative results, we also compare qualitative results of baseline approaches and our methods in Fig.~\ref{fig:qual_examples}. For all the samples, we can observe that our methods (\ourmodelDepth~and \ourmodel) achieve the best counting results with excellent object localizations.

\myPara{Complexity analysis.} Though \ourmodelDepth~clearly outperforms \ourmodel~in terms of all metrics (MAE, MSE, and NAE) due to the exploitation of depth modality, it only increases the number of parameters from 52.5M to 59.4M. The reason is that we use small depth decoder and efficient agent attention.

\begin{table}[!t]
\centering
\caption{\textbf{Impact of the number of transformer layers or convolutional blocks in DETE.} For our default setting, we choose $L=2$ due to good balance between performance and model complexity.}
\label{table:ablation_l}
% \footnotesize
\small
\renewcommand{\arraystretch}{1.35}
\renewcommand{\tabcolsep}{7.0mm}
%\setlength{\tabcolsep}{8pt}
%\resizebox{0.48\textwidth}{!}{%
\begin{tabular}{c|c|c|c}
\hline
$L$ & MAE$\downarrow$   & MSE$\downarrow$   & NAE$\downarrow$  \\ \hline
2 & 16.75 & 35.87 & 0.28 \\ 
4 & 16.59 & 35.23 & 0.26 \\ 
6 & 15.91 & 34.08 & 0.26 \\ 
8 & \textbf{15.72} & \textbf{33.63} & \textbf{0.24} \\ \hline
\end{tabular}
%}
\end{table}

\begin{table}[!t]
\centering
\caption{\textbf{Impact of $L_m$, the number of feature interactions between between RGB and depth modalities in CAMF.} For our default setting, we choose $L_m=2$.}
\label{table:ablation_lm}
\small
\renewcommand{\arraystretch}{1.35}
\renewcommand{\tabcolsep}{6.0mm}
\begin{tabular}{c|c|c|c}
\hline
$L_m$ & MAE$\downarrow$   & MSE$\downarrow$   & Params (M)  \\ \hline
0 & 15.78 & 34.2 & 55.9 \\ 
1 & 15.56 & 34.37 & 57.7 \\ 
2 & \textbf{15.19} & \textbf{32.89} & 59.4 \\ 
3 & 15.43 & 33.03 & 61.1 \\ \hline
\end{tabular}
\end{table}

\subsection{Ablation Study}
% \vspace{-3pt}
\myPara{Impact of the density branch and DETE.} As mentioned, the proposed model combines a density and a regression branch in a unified framework, aiming to combine their advantages. In Table~\ref{table:ablation_structure}, we show the results of separately training the density branch and the regression branch. We also provide results of jointly training the density branch and regression branch without using the proposed DETE. The comparison shows that the regression branch, though straightforward, performs better than only using the density branch.
Furthermore, training both branches together without DETE gives better performance than using only the regression branch. The improvement could be explained from the perspective of multi-task learning~\cite{caruana1997multitask,weinberger2009feature,sun2021task}. The added density branch, which could be regarded as an \textit{additional task}, helps the encoder learn better features. By establishing connections between the density and regression branches, better performance is obtained. Compared to the variant without DETE, our final model has a clear superiority by reducing MAE from 16.94 to 15.91 and MSE from 35.92 to 34.08. The results validate the effectiveness of DETE for enhancing the features by exploiting the information generated from the density branch.

\myPara{Impact of $L$.} We change the number of $\rm Trans$ or $\rm Convs$ in DETE and report results in Table~\ref{table:ablation_l}. By increasing $L$, we obtain better performance, showing the capability of our DETE to produce relevant features. We use $L=6$ in our main setting to balance complexity and performance.

\myPara{Impact of $L_m$ and effectiveness of CAMF module.} We study the impact of $L_m$, the number of modality interactions in cross-agent multimodal fusion (CAMF) module. The detailed results are shown in Table~\ref{table:ablation_lm}. When $L_m=0$, we directly add RGB and depth features to generate $F_m$, which serves as a baseline. It can be observed that our design of adopting cross-agent attention is effective in exploiting the complementary information within different modalities since using cross-agent attention always outperforms the version without using it ($L_m=0$). When increasing $L_m$ from 1 to 3, we found 2 times of feature interactions leads to the best performance, giving MAE and MSE of 15.19 and 32.89, respectively. Also, large $L_m$ leads to high number of parameters. The model ($L_m=1$) has 57.5M parameters while the one using 3 times of modality interactions has 61.1M parameters. Therefore, when setting $L_m=2$, an excellent balance between performance and model complexity can be achieved.

\section{Conclusions and Future Work}
% \vspace{-4pt}
We provide a rigorous study of a new challenge named indiscernible object counting (IOC), which focuses on counting objects in indiscernible scenes. To address the lack of a large-scale dataset, we present the high-quality \ourdataset~which mainly contains underwater scenes and has point annotations located at the center of object (mainly fish) instances. A number of existing mainstream baselines are selected and evaluated on \ourdataset, proving a domain gap between DOC and IOC. In addition, we propose a dedicated method for IOC named \ourmodel, which is equipped with two novel designs: combining a density and regression branch in a unified model and a density-enhanced transformer encoder which transfers object density information from the density to the regression branch. \ourmodel~achieves SOTA performance on \ourdataset. 

To unleash the power of multimodal information in IOC, we extend \ourdataset~to \ourdatasetDepth, which now contains high-quality depth maps for all images. For benchmarking purposes, we evaluate 4 existing state-of-the-art multimodal counting methods on \ourdatasetDepth. We also extend \ourmodel~to \ourmodelDepth~by equipping it with the ability to effectively integrate useful information from different modalities through the proposed cross-agent multimodal fusion module.

To sum up, our datasets (\ourdataset~and \ourdatasetDepth) and methods (\ourmodel~and \ourmodelDepth) provide an opportunity for future researchers to dive into this new task.

\textbf{Future work.}
There are several directions. (1) To improve performance and efficiency. Although our method achieves state-of-the-art performance, there is room to further improve the counting results on \ourdataset~and \ourdatasetDepth~in terms of MAE, MSE, and NAE. Also, efficiency is an important factor when deploying counting models in real applications. (2) To study domain adaptation among IOC and DOC. There are many more DOC datasets than IOC datasets and how to improve IOC using available DOC datasets is a practical problem to tackle. (3) To obtain a general counting model which can count everything (people, plants, cells, fish, \etc). 

% \vspace{-6pt}
\section*{Acknowledgement}
% \vspace{-4pt}
This work is supported in part by Toyota Motor Europe through a research project TRACE-Zürich and in part by NSFC (No. 62476143). We would like to thank Dr. Ce Liu and Dr. Christos Sakaridis for their valuable suggestions. 

{\small
\bibliographystyle{IEEEtran}
\bibliography{egbib}
}

\newcommand{\AddPhoto}[1]{\includegraphics[width=1in,keepaspectratio]{Authors/#1}}

\begin{IEEEbiography}[\AddPhoto{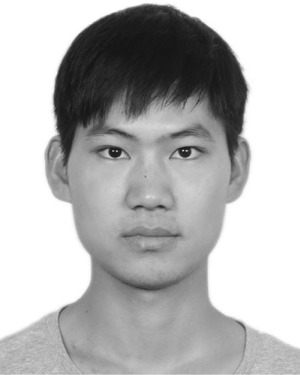}]{Guolei Sun}
received his Ph.D. degree at ETH Zurich, Switzerland, in Prof. Luc Van Gool's Computer Vision Lab in Jan 2024.  Before that, he got master degree in computer science from the King Abdullah University of Science and Technology (KAUST), in 2018. He is currently a postdoctoral researcher at Computer Vision Lab, ETH Zurich. From 2018 to 2019, he worked as a research engineer with the Inception Institute of Artificial Intelligence, UAE. His research interests include deep learning for video understanding, semantic/instance segmentation, object counting, and weakly supervised learning.
\end{IEEEbiography}

\begin{IEEEbiography}[\AddPhoto{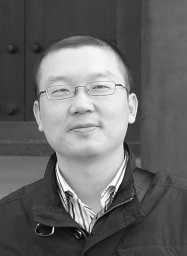}]{Xiaogang Cheng}
is an Associate Professor at Nanjing University of Posts and Telecommunications (NUPT). He obtained his degrees from Nanjing University and Southeast University. His research interests include: 1) vision-based non-invasive perception for human thermal comfort, 2) foggy and hazy visibility perception, Reflection removal.
\end{IEEEbiography}

\begin{IEEEbiography}[\AddPhoto{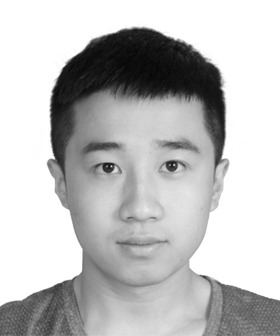}]{Zhaochong An}
is a PhD student at the University of Copenhagen, affiliated with Pioneer Centre for Artificial Intelligence, under the ELLlS program. He is advised by Prof. Serge Belongie. He received his MSc in Computer Science from ETH Zürich in 2023 under the supervision of Prof. Luc Van Gool. His research primarily focuses on computer vision and deep learning, including scene understanding, video analysis, few-shot/unsupervised learning, and multimodal learning.
\end{IEEEbiography}

\begin{IEEEbiography}[\AddPhoto{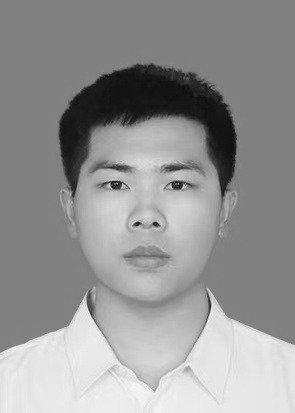}]{Xiaokang Wang}
is a master student at Nanjing University of Posts and Telecommunications (NUPT). His research interest includes computer vision and AI for science, covering areas of scene understanding, object counting, and image/video segmentation.
\end{IEEEbiography}

\begin{IEEEbiography}[\AddPhoto{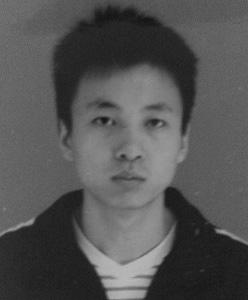}]{Yun Liu}
received his B.E. and Ph.D. degrees from Nankai University in 2016 and 2020, respectively. Then, he worked with Prof. Luc Van Gool as a postdoctoral scholar at Computer Vision Lab, ETH Zurich, Switzerland. After that, he worked as a senior scientist at the Institute for Infocomm Research (I2R), A*STAR, Singapore. Currently, he is a professor at the College of Computer Science, Nankai University. His research interests include computer vision and machine learning.
\end{IEEEbiography}

\begin{IEEEbiography}[\AddPhoto{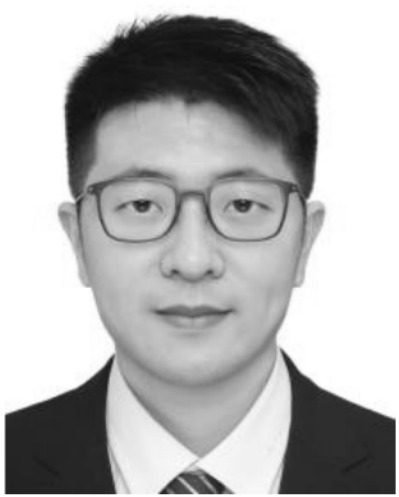}]{Deng-Ping Fan}
is a Full Professor and Deputy Director of the Media Computing Lab (MCLab) in the College of Computer Science at Nankai University, China. His research interests span computer vision, machine learning, and medical image analysis.
\end{IEEEbiography}

\begin{IEEEbiography}[\AddPhoto{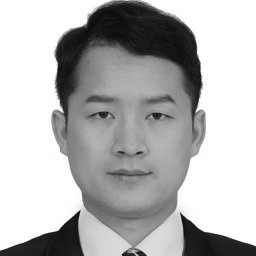}]{Ming-Ming Cheng}
(Senior Member, IEEE) received the PhD degree from Tsinghua University, in 2012. Then he did 2 years research fellow, with Prof. Philip Torr in Oxford. He is now a professor with Nankai University, leading the Media Computing Lab. His research interests include computer graphics, computer vision, and image processing. He received research awards including ACM China Rising Star Award, IBM Global SUR Award, CCF-Intel Young Faculty Researcher Program.
\end{IEEEbiography}

\begin{IEEEbiography}[\AddPhoto{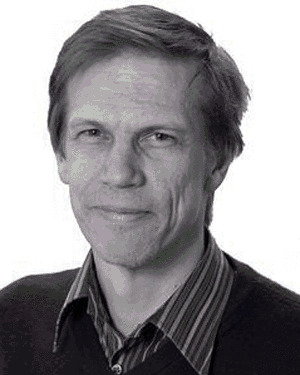}]{Luc Van Gool}
received the degree in electromechanical engineering from the Katholieke Universiteit Leuven, in 1981. Currently, he is a professor with the Katholieke Universiteit Leuven in Belgium and the ETH in Zurich, Switzerland. He leads computer vision research with both places, and also teaches at both. He has been a program committee member of several major computer vision conferences. His main research interests include 3D reconstruction and modeling, object recognition, tracking, gesture analysis, and the combination of those. He received several Best Paper awards, won a David Marr Prize and a Koenderink Award, and was nominated Distinguished Researcher by the IEEE Computer Science committee. He is a co-founder of 12 spin-off companies.
\end{IEEEbiography}

\vfill

\end{document}